\documentclass[letterpaper]{article}
\let\articleaddcontentsline\addcontentsline

\usepackage[preprint]{aaai2027}

\usepackage[hyphens]{url}
\usepackage{graphicx}
\usepackage{natbib}
\usepackage{caption}
\usepackage{booktabs}
\usepackage{subfiles}
\usepackage{tabularx}
\usepackage[table]{xcolor}
\usepackage{amsmath}
\usepackage{amssymb}
\usepackage{array}
\usepackage{multirow}
\usepackage{adjustbox}
\usepackage{pifont}
\usepackage{float}
\usepackage{framed}
\usepackage{fvextra}

\graphicspath{{./Figures/}}

\newcommand{\code}[1]{\texttt{\detokenize{#1}}}
\newcommand{\BasketballBench}{\emph{BasketballBench}\ }
\newcommand{\BasketballSkills}{\emph{BasketballSkills}\ }

\DefineVerbatimEnvironment{PromptVerbatim}{Verbatim}{
  fontsize=\footnotesize,
  breaklines=true,
  breakanywhere=true,
  breaksymbolleft={},
  breaksymbolright={},
  obeytabs=true,
  tabsize=2
}

\title{Towards Comprehensive Basketball Understanding}

\author{
Yirong Hu\textsuperscript{\rm 1},
Jiayuan Rao\textsuperscript{\rm 2},
Yu Zhang\textsuperscript{\rm 2},
Shangzhe Di\textsuperscript{\rm 2},
Weidi Xie\textsuperscript{\rm 2}
}

\affiliations{
\textsuperscript{\rm 1}School of Mathematical Sciences,
Peking University, China\\
\textsuperscript{\rm 2}School of Artificial Intelligence,
Shanghai Jiao Tong University, China\\
huyirong@stu.pku.edu.cn,
\{jy\_rao, zhangyu2012, weidi\}@sjtu.edu.cn,
shangzhe.di@gmail.com
}
\begin{document}
\maketitle
\setlength{\parskip}{0pt}
\begin{abstract}
Understanding a basketball game requires recognizing events, localizing actions, identifying players, and relating these to structured game knowledge.
Existing benchmarks primarily evaluate these abilities one at a time, leaving the interactions among these abilities under-explored. We introduce \textbf{BasketballBench}, a multimodal benchmark comprising 7,980 questions across ten tasks in text, image, and video.
It is built from the 2025--2026 NBA season and includes official play-by-play, rosters and profiles for 530 active players, and 2,501 possession-level broadcast clips.
We further propose \textbf{BasketballSkills}, an agent that composes eight basketball-specific perception and retrieval tools under four reusable skills that specify tool order, evidence bindings, and stopping conditions. 
Experiments show that current MLLMs struggle particularly on questions requiring the integration of multiple capabilities, whereas BasketballSkills outperforms them, highlighting the effectiveness of explicitly composing domain-specific capabilities for comprehensive basketball understanding.
\end{abstract}

\vspace{6pt}

\section{Introduction}
Multimodal large language models (MLLMs) have made substantial progress in visual recognition and question answering, yet reliable understanding of professional basketball remains challenging. 
Understanding a single basketball play may require a model to recognize the event, identify the relevant players, localize the action in space and time, interpret the broadcast context, and connect visual evidence with structured basketball knowledge. 
We refer to the integration of these heterogeneous sources of evidence as comprehensive basketball understanding.

\vspace{4pt}

Research on sports understanding has addressed a broad spectrum of capabilities, ranging from player- and event-level perception to tasks such as professional QA, rule-aware reasoning, long-form temporal analysis, and multi-view understanding~\cite{ramanathan2016detecting,li2021multisports,cui2023sportsmot,meng2026soccerref,li2024sportsqa,rao2024matchtimeautomaticsoccergame,xia2025sportu,xia2026sportr,cao2026sportstime,chen2026sportmv,pan2025basket}. 
Existing works still tend to evaluate these capabilities separately, often using different annotation schemes, output formats, and sports, while some recent work has begun to bring multiple capabilities together within a unified framework~\cite{rao2025soccceragent}.
This motivates a basketball-specific benchmark for evaluating their composition and a structured framework for organizing them into reusable procedures.

\vspace{4pt}

To provide a unified evaluation setting, we construct \textbf{BasketballBench}, a multimodal benchmark comprising 7,980 annotated instances across ten basketball-specific tasks, examining basketball understanding from three perspectives: (1) basketball knowledge and retrieval, (2) broadcast and player perception, and (3) spatiotemporal and event understanding. The ten tasks span text, image, and video inputs and range from single-capability tests to composite questions so that the contributing capabilities can be read off separately.
This organization supports consistent model comparison and fine-grained error diagnosis, while revealing a recurring set of basketball-specific capabilities that can be further formalized and composed as reusable skills.

\vspace{4pt}

To turn these recurring capabilities into executable components, we introduce \textbf{BasketballSkills}, a hierarchical framework comprising eight atomic tools and four reusable procedural skills.
The tools perform basketball-specific perception and retrieval operations, while the skills organize them into multi-step workflows dynamically selected and executed by a language-model controller, with each tool call validated by a lightweight verifier.
Together, they support all ten benchmark tasks without predefined task labels or fixed pipelines, while keeping intermediate evidence inspectable.

\vspace{4pt}

In summary, we make \textbf{three contribution} in this work:
(i) we introduce BasketballBench, a comprehensive multimodal benchmark that evaluates basketball understanding through ten tasks covering both individual capabilities and their composition;
(ii) we present BasketballSkills, a unified framework that represents basketball-specific capabilities as reusable and composable skills and dynamically selects them according to the question and available evidence;
(iii) through systematic experiments across all benchmark tasks, we show that BasketballSkills outperforms the best commercial MLLM on eight of the ten tasks.
Together, these contributions establish a unified foundation for evaluating, diagnosing, and advancing knowledge-grounded and skill-composable sports understanding.

\section{Related Work}

\noindent \textbf{Sports understanding and evaluation.}
Sports understanding has been studied across multiple levels of analysis, from player- and event-level perception~\cite{giancola2018soccernet1,deliege2021soccernet2,rao2025unisoccer} to structured game interpretation. 
Representative work has examined event and key-actor localization~\cite{ramanathan2016detecting}, 
dense spatiotemporal action detection~\cite{li2021multisports}, 
multi-object tracking under fast motion and similar player appearance~\cite{cui2023sportsmot}, and jersey-number recognition~\cite{koshkina2024jersey}. 
Within basketball, specialized studies have further addressed court calibration~\cite{sha2020calibration}, 
player identification~\cite{senocak2018player}, 
knowledge-enhanced description~\cite{wang2021pdvc,xi2025keanet}
and fine-grained player evaluation~\cite{pan2025basket}. 
These efforts provide important perceptual and structural components, but they are generally developed and evaluated as separate problems.
More recent benchmarks have extended sports evaluation toward professional question answering and multimodal reasoning. Sports-QA studies descriptions, temporal relations, causality, and counterfactual reasoning~\cite{li2024sportsqa}; SPORTU and SportR examine rule- and strategy-aware reasoning~\cite{xia2025sportu,xia2026sportr}; and SportsTime and SportMV-Bench expand evaluation to long-form temporal evidence and multiple camera views~\cite{cao2026sportstime,chen2026sportmv}. 
Despite this broader coverage, these benchmarks provide limited evaluation of how capabilities interact.

\vspace{5pt}\noindent \textbf{Skills for multimodal reasoning.}
Complex multimodal problems are increasingly addressed through modular reasoning, in which language models decompose a request and interact with external operations. ReAct interleaves reasoning with actions and observations~\cite{yao2023react}, while VisProg and ProViQ translate natural-language requests into executable visual or video programs~\cite{gupta2023visprog,choudhury2024proviq}. Beyond task-specific programs, Voyager and MMSkills explore reusable skill representations that allow procedural knowledge to be stored, adapted, and composed across tasks~\cite{wang2024voyager,zhang2026mmskills}. In sports, SportMV-Agent applies iterative planning and evidence collection to multi-view reasoning~\cite{chen2026sportmv}.

Existing modular and skill-based systems, however, are mainly designed for general-purpose environments or narrowly defined reasoning settings. They do not explicitly organize the capabilities of a professional sport into a shared set of domain-grounded, executable skills that can also be evaluated systematically. Therefore, we built BasketballSkills that extend this line of research by representing basketball-specific capabilities as reusable skills with explicit inputs, outputs, and dependencies.

\section{BasketballBench}
\label{sec:basketballbench}

This section first presents the benchmark overview in Section~\ref{subsec:basketballbench-motivation}, then describes its data sources and tasks in Sections~\ref{subsec:basketballbench-datasources} and~\ref{subsec:basketballbench:benchmark_tasks}, and finally summarizes the construction procedure in Section~\ref{subsec:basketballbench:construction}.

\subsection{Overview}
\label{subsec:basketballbench-motivation}
Comprehensive basketball understanding requires models to demonstrate several complementary capabilities, including domain knowledge, visual perception, spatial-temporal understanding, and event-level reasoning. 
These aspects examine whether a model can retrieve basketball facts, recognize players and broadcast information, understand where and when actions occur, and recover structured events from game footage. 
Based on this capability taxonomy, \textbf{BasketballBench} contains 7,980 QA pairs across ten tasks with text, image, and video inputs. Unlike most sports benchmarks, BasketballBench links visual events to spatial locations, 
timestamps, participants, on-court identities, and structured basketball knowledge at the instance level.

\begin{figure*}[t] 
    \centering     
    \includegraphics[width=\textwidth]{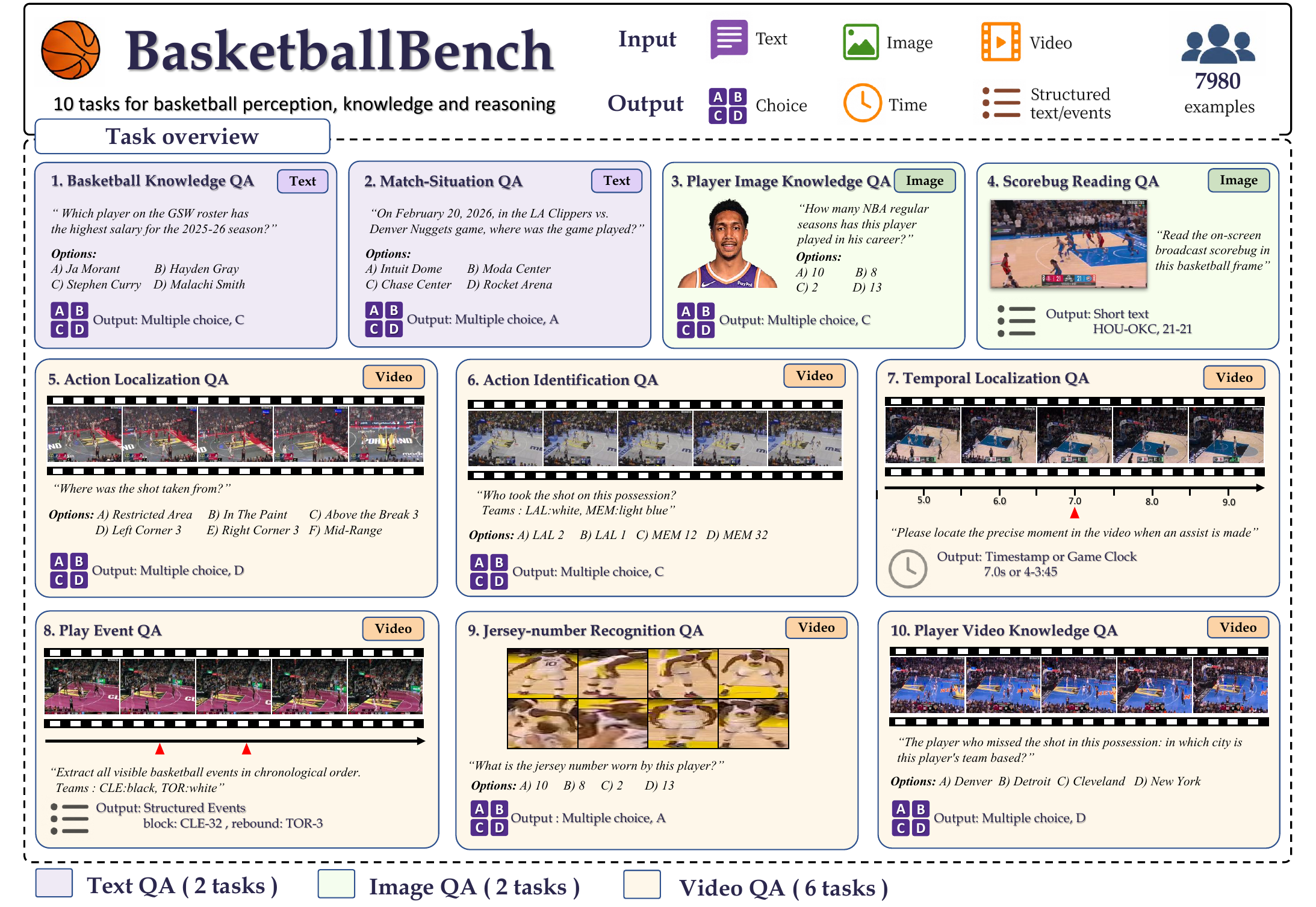}

    \vspace{-12pt}
    \caption{Representative examples of the ten BasketballBench tasks. The benchmark covers different aspects with outputs ranging from multiple-choice answers and timestamps to structured event sequences.}
    \vspace{-7pt}
    
    \label{fig:basketballbench} 
\end{figure*}

\subsection{Data Sources}\label{subsec:basketballbench-datasources}

BasketballBench is built from three complementary data sources
from the 2025--2026 NBA season.

\paragraph{Broadcast Videos.}
We select 2,501 possession-level broadcast clips following the
test split of BasketEvent~\cite{zhang2026basketeventunderstandingdidbasketball}. Each clip is
associated with its source game and aligned with the corresponding
structured play-by-play records.

\paragraph{Structured Records.}
We collect schedules, official play-by-play records, game officials,
team metadata, and game-specific uniform assignments from
NBA.com~\cite{nba_official} and Sportradar~\cite{nba_api}.
We additionally collect the rosters and coaching staffs of all
30 teams and profiles for 530 active players. These records are
normalized into a relational SQL database that supports reproducible
question generation and answer retrieval.

\paragraph{Player Images.}
We collect two complementary image sets for the 530 active players.
First, we obtain one official headshot per player from NBA.com,
forming a canonical reference collection linked to the player
profiles in our SQL database. Second, we manually collect 400
additional photographs for use as the visual inputs to the
player-identification examples. Each additional photograph is
manually verified to depict the associated player and to be different
from that player's official headshot. The identity links are retained
only as benchmark annotations and are not exposed as part of the
visual input. This separation ensures that the evaluated systems must
identify players across different images rather than retrieve an
identical reference photograph.

\subsection{Benchmark Tasks}
\label{subsec:basketballbench:benchmark_tasks}
BasketballBench comprises ten tasks spanning different input modalities, answer formats, and levels of reasoning complexity. 
To organize this diversity in a coherent manner, we group the tasks into three categories based on the types of evidence they use and the reasoning required to derive an answer. 
Table~\ref{tab:basketballbench_tasks} summarizes the main settings of each task, while Figure~\ref{fig:basketballbench} provides representative examples.
\paragraph{Basketball Knowledge and Retrieval.}
This group evaluates basketball knowledge under different forms of knowledge context.
(1)~\textit{Basketball Knowledge QA} covers general facts about players,
teams, schedules, and statistics.
(2)~\textit{Match-situation QA} focuses on information associated with a specific game.
(3)~\textit{Player Image Knowledge QA} \& (4)~\textit{Player  Video Knowledge QA} combine player recognition with knowledge-based question answering that the model must identify the relevant player and then retrieve further information.

\begin{table*}[t]
\centering
\small
\renewcommand{\arraystretch}{1.25}          
\rowcolors{2}{gray!10}{white}               
\begin{tabularx}{\textwidth}{@{} 
    l 
    l
    >{\centering\arraybackslash}p{2.2cm} 
    r 
    >{\raggedright\arraybackslash}p{2.8cm} 
    >{\raggedright\arraybackslash}X 
@{}}
\toprule
\textbf{Index} & \textbf{Task} & \textbf{Input} & \textbf{Samples} & \textbf{Output} & \textbf{Primary capability} \\
\midrule
 Q1 & Basketball Knowledge QA        & Text        & 1,200 & Multiple Choice   & Domain Knowledge Grounding \\
 Q2 & Match-Situation QA             & Text        & 1,200 & Multiple Choice   & Contextual Retrieval \\
 Q3 & Player Image Knowledge QA      & Image       & 400   & Multiple Choice   & Domain Knowledge Grounding \\
 Q4 & Scorebug Reading QA            & Image       & 200   & Structured Text   & Broadcast Understanding \\
 Q5 & Action Localization QA         & Video       & 600   & Multiple Choice   & Spatial Understanding \\
 Q6 & Action Identification QA       & Video       & 500   & Multiple Choice   & Player Identification\\
 Q7 & Temporal Localization QA          & Video       & 1,580 & Timestamp         & Temporal Understanding \\
 Q8 & Play Event QA                  & Video       & 1,000 & Structured Events & Event Understanding\\
 Q9 & Jersey-Number Recognition QA   & Video       & 300   & Multiple Choice   & Broadcast Understanding \\
 Q10 & Player Video Knowledge QA     & Video       & 1,000 & Multiple Choice   & Domain Knowledge Grounding \\
\bottomrule
\end{tabularx}
\vspace{-6pt}
\caption{BasketballBench comprises 7,980 examples across 10 tasks, covering text, image, and video inputs, and assesses basic perceptual abilities, basketball domain knowledge, and reasoning.
}
\vspace{-12pt}
\label{tab:basketballbench_tasks}
\end{table*}

\paragraph{Broadcast and Player Perception.}
This group focuses on information that must be extracted directly from basketball broadcasts. 
(1)~\textit{Scorebug Reading QA} consists of two modes: recovering the current teams and score, and recovering the game clock information from broadcast graphics.
(2)~\textit{Jersey-Number Recognition QA} aggregates evidence across multiple frames to identify a partially visible player number, while 
(3)~\textit{Action Identification QA} associates an observed basketball action (\textit{e.g.} shot) with its corresponding player identity.
Together, these tasks cover broadcast-level, player-level, and action-level visual evidence.

\paragraph{Spatiotemporal and Event Understanding.}
The remaining tasks examine how basketball actions are situated and organized within a possession. 
(1)~\textit{Action Localization QA} determines the spatial region from which an action is finished, and 
(2)~\textit{Temporal Localization QA} localizes a target action in either video time or game-clock coordinates. 
(3)~\textit{Play Event QA} moves beyond a single prediction by recovering an ordered sequence of visible events and their participants. 
These tasks therefore progress from spatial and temporal localization to structured interpretation of a complete possession.

\subsection{Benchmark Construction}
\label{subsec:basketballbench:construction}

We construct the ten tasks by instantiating manually designed QA templates from structured records and pairing them with player photographs or event-aligned clips when needed. Scorebug annotations combine source-game metadata with initial predictions from Qwen3.6-27B~\cite{qwen2026qwen36}, while SAM 3~\cite{carion2025sam3}-derived trajectories support jersey annotation; all model-assisted labels are manually verified. Aligned events, participants, game clocks, and shot locations are converted into task-specific targets, with shots mapped to six court regions and participants represented by game-specific color--number pairs. Further construction details are provided in the \textbf{Supplementary Materials}.

\paragraph{Human quality control.}
The amount of human verification depends on the provenance and
difficulty of each task. Q1 and Q2 are generated directly from
structured official database records and do not receive additional
instance-level human review. For Q3, the official gallery headshots
are not separately audited; however, all 400 evaluation photographs
are manually collected and checked to ensure that each image depicts
the corresponding player and is different from that player's official
headshot. For Q4, Qwen3.6-27B produces 300 candidate examples, from
which human reviewers select 200 correct and readable examples for
inclusion in the benchmark, corresponding to a retention rate of
66.7\%. Q5 and Q6 are generated from structured play-by-play, shot,
roster, and game metadata without additional instance-level human
review.

Q7 receives full manual timestamp annotation because timestamps in
official play-by-play records can lag the visible occurrence of an
event by several seconds. For each of the 790 target event instances,
the clip-relative video timestamp and the corresponding game-clock
time are manually reannotated. Each annotation is subsequently
converted into two questions, resulting in 1,580 Q7 questions. All
1,000 Q8 examples are manually checked for event completeness,
temporal order, event type, team, and participant jersey number. This
review corrects 258 examples (25.8\%), while the remaining 742 are
verified without modification. For Q9, Qwen3.5-27B produces the
initial candidates, all of which are manually reviewed; only correct
and unambiguous candidates are retained, yielding 300 final examples.
Q10 is generated from aligned play-by-play and player metadata without
additional instance-level human review. For tasks without instance-level
human review, we still apply deterministic schema validation,
missing-field filtering, and duplicate removal during construction.

\section{Methodology}
\label{sec:methodology}
This section presents \textbf{BasketballSkills}, a hierarchical framework for answering basketball questions. 
We first formulate the problem and clarify the roles of queries, tools, skills, and the controller in Section~\ref{subsec:methodology:problem}. 
Then in Section~\ref{subsec:methodology:architecture}, we introduce the architecture, including the two-level libraries and the on-demand skill-loading mechanism. 
Finally, Section~\ref{subsec:methodology:workflow} presents the workflow of BasketballSkills.

\begin{figure*}[t] 
    \centering     
    \includegraphics[width=\textwidth]{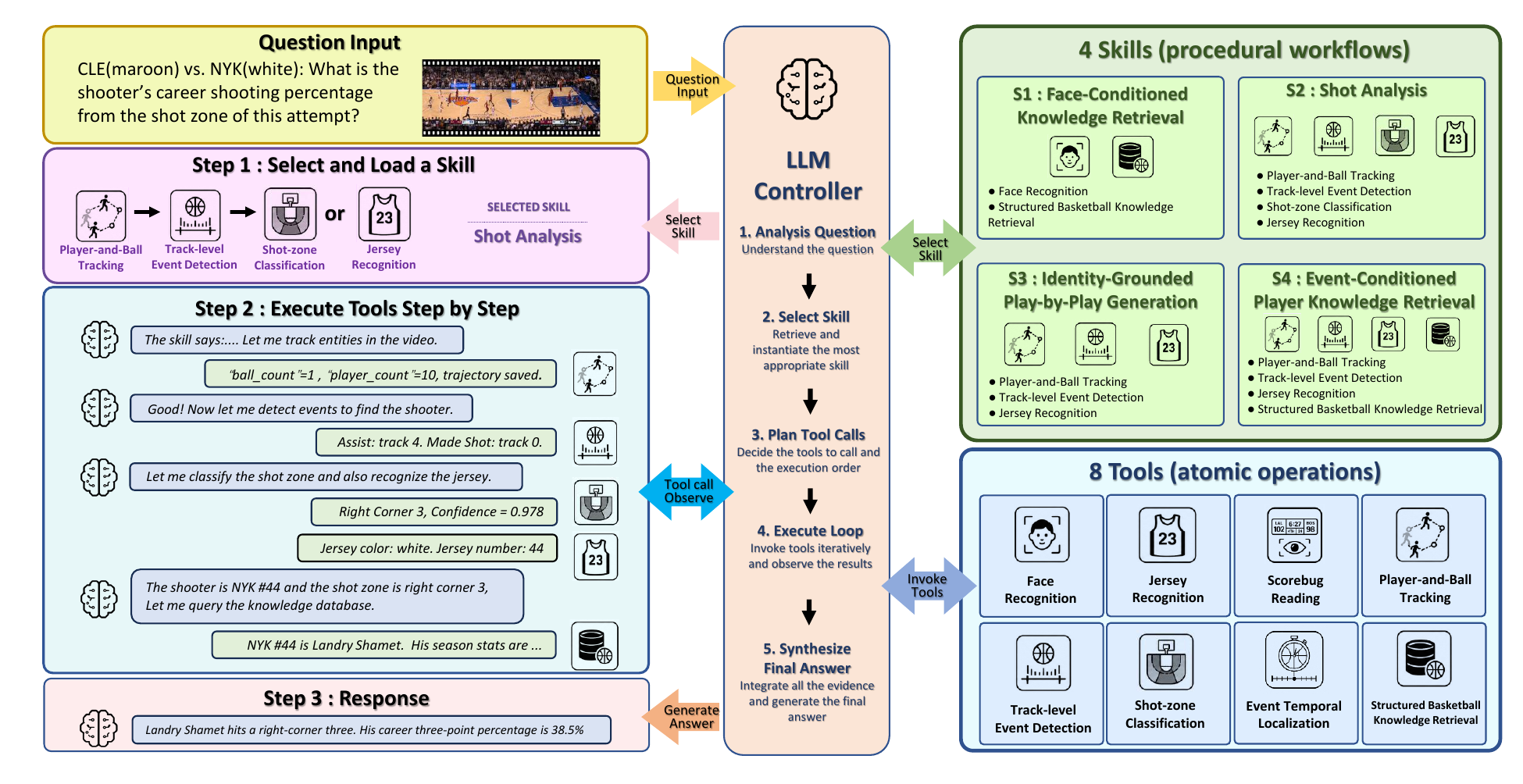}
    \vspace{-15pt}
    \caption{BasketballSkills Architecture Overview. BasketballSkills selectively loads the appropriate skills for each task and, following their guidance, composes and invokes specialized tools to complete the task.}
    \vspace{-10pt}
    
    \label{fig:basketballskills_architecture}
\end{figure*}
\subsection{Problem Formulation}
\label{subsec:methodology:problem}

We formulate BasketballSkills as a framework for solving these heterogeneous questions through reusable basketball-specific skills with atomic tools.
Let \(x=(q,m)\) denote a multimodal basketball query, where \(q\) is a textual query and \(m\) represents context, including images, videos, and game metadata. 
BasketballSkills consists of three components: 
(1) an atomic tool library \(\mathcal{T}=\{\tau_1,\ldots,\tau_N\}\), in which each tool performs a single, clearly defined operation; 
(2) a skill library \(\mathcal{S}=\{s_1,\ldots,s_M\}\), where each skill \(s_k\) organizes multiple tools into a reusable procedure;
and (3) language-model controller \(\pi_{\theta}\) that interprets the query, selects the skills, and coordinates the corresponding executions.
Formally, each tool is a typed operator $\tau_j:\mathcal{X}_j\rightarrow\mathcal{O}_j$,
where \(\mathcal{X}_j\) and \(\mathcal{O}_j\) denote its input and output spaces. 
When invoked with arguments \(u_i\), a tool returns a structured result
\begin{equation}
    o_i=\tau_{j_i}(u_i),
\end{equation}
which may be used by the controller or passed to subsequent tools. Tool invocation is optional and dynamically determined by the query and intermediate results.
Denote \(\mathcal{F}_{\pi_\theta}\) as the complete inference process coordinated by the controller, the overall objective is to generate an answer \(y\) as
\begin{equation}
y=\mathcal{F}_{\pi_\theta}(x;\mathcal{S},\mathcal{T}),    
\end{equation}

\subsection{Skills Architecture}
\label{subsec:methodology:architecture}

BasketballSkills adopts a two-level architecture consisting of atomic tools and  composite skills. 
As shown in Figure~\ref{fig:basketballskills_architecture}, tools provide responses through standardized interfaces, while skills organize multiple tools into reusable workflows, with all the details in \textbf{Supplementary Materials}.

\paragraph{Tool Library.}
Our atomic tool library $\mathcal{T}$ comprises 8 tools spanning visual perception and structured basketball knowledge. 
Specifically, it includes seven perception tools: (1)~\textit{face recognition}, (2)~\textit{jersey recognition}, (3)~\textit{scorebug reading}, (4)~\textit{player-and-ball tracking},  (5)~\textit{track-level event detection}, (6)~\textit{shot-zone classification}, and (7)~\textit{event temporal localization}; 
and one knowledge tool: (8)~\textit{structured basketball knowledge retrieval}, which queries records of players, teams, games, and statistics. 
Each tool exposes typed inputs and structured outputs, allowing evidence produced by one tool to be consumed by subsequent operations.

\paragraph{Tool Implementations.}
All tools are implemented using open-source models. Here, we introduce the basic implementations of different tool groups.

\vspace{0.06cm}
\noindent\textit{(1)~Face recognition.}
We use the open-source face recognition library~\cite{geitgey2017facerecognition} to encode an input face and match it against the gallery of official player headshots.

\vspace{0.06cm}
\noindent\textit{(2)~Prompted visual reading.}
The tools of scorebug reading, jersey recognition, and event temporal localization share a Qwen3.5-9B~\cite{qwen2026qwen35} backend. Each tool receives its supported visual materials with a task-specific prompt and returns a typed output, such as scorebug fields, jersey attributes, or an event timestamp.

\vspace{0.06cm}
\noindent\textit{(3)~Player-and-ball tracking.}
We combine SAM 3~\cite{carion2025sam3} with a fine-tuned
RF-DETR~\cite{robinson2026rfdetr} for player-and-ball tracking,
using RF-DETR detections to filter erroneous SAM trajectories.

\vspace{0.06cm}
\noindent\textit{(4)~Event and shot-zone models.}
We use PlayNet~\cite{zhang2026basketeventunderstandingdidbasketball}
to recognize event types and associate them with participant tracks.
For shot-zone classification, we add a six-class classification head
to PlayNet and fine-tune it to predict the court region of the target shot.

\vspace{0.06cm}
\noindent\textit{(5)~Structured basketball knowledge retrieval.}
This tool exposes fixed query interfaces over the SQL database described in Section~\ref{subsec:basketballbench-datasources}, including player, roster, schedule, game, and statistical records, with a controller determining the invocation order and composing their responses with perceptual evidence.

\paragraph{Skill Library.}
The composite skill library $\mathcal{S}$ contains four reusable basketball workflows built upon the atomic tool library $\mathcal{T}$. Each skill specifies tool ordering, intermediate evidence flow, and task-specific output construction.

\vspace{0.1cm}
\noindent\textit{(1) Face-Conditioned Knowledge Retrieval.}
This skill invokes the Face Recognition and Structured Basketball Knowledge Retrieval tool to identify a player from a face image to the corresponding profile information.

\vspace{0.1cm}
\noindent\textit{(2) Shot Analysis.}
This skill combines  Player-and-Ball Tracking with Track-Level Event Detection to locate a target shot and its shooter track. It then invokes Shot-Zone Classification or Jersey Recognition according to the query, using the provided team-color mapping to resolve the shooter's on-court identity when needed.

\vspace{0.1cm}
\noindent\textit{(3) Identity-Grounded Play-by-Play Generation}
This skill uses Player-and-Ball Tracking and Track-Level Event Detection to recover ordered events and their participant tracks. Jersey Recognition then grounds participating tracks to team and jersey-number identities, which are serialized into a chronological play-by-play record.

\vspace{0.1cm}
\noindent\textit{(4) Event-Conditioned Player Knowledge Retrieval.}
This skill locates the participant of a specified video event through tracking and event detection, resolves the player from jersey attributes and team context, and retrieves the requested information through Structured Basketball Knowledge Retrieval.

\subsection{Inference Workflow}
\label{subsec:methodology:workflow}

At inference time, the DeepSeek-V4-Flash controller~\cite{deepseekv4} receives the multimodal query $x$, the atomic tool library $\mathcal{T}$, and a lightweight catalog of the skill library $\mathcal{S}$. 
It invokes a relevant skill in the workflow as:
\begin{equation}
    s_k = \operatorname{Select}(x,\mathcal{S}),
    \qquad
    y = \pi_{\theta}(x;\operatorname{load}(s_k),\mathcal{T}).
\end{equation}
The controller then invokes tools according to the loaded skill and adapts subsequent calls based on their observations. Each call is validated by a lightweight verifier, and the final answer is generated once sufficient evidence has been collected.

\begin{table*}[t]
\centering
\footnotesize
\renewcommand{\arraystretch}{1.08}
\setlength{\tabcolsep}{4.0pt}

\begin{adjustbox}{max width=\textwidth,center}
\begin{tabular}{
    @{}l
    |cc
    |cc
    |cccccc
    |ccc@{}
}
    \toprule[1.2pt]

    \multirow{2}{*}{\textbf{Model}}
    & \multicolumn{2}{c}{\textbf{TextQA}}
    & \multicolumn{2}{c}{\textbf{ImageQA}}
    & \multicolumn{6}{c}{\textbf{VideoQA}}
    & \multicolumn{3}{c}{\textbf{Overall}}
    \\

    \cmidrule(lr){2-3}
    \cmidrule(lr){4-5}
    \cmidrule(lr){6-11}
    \cmidrule(lr){12-14}

    & Q1 & Q2
    & Q3 & Q4
    & Q5 & Q6 & Q7 & Q8 & Q9 & Q10
    & Text & Image & Video
    \\

    \midrule[0.7pt]
    \multicolumn{14}{c}{\textit{Commercial APIs}} \\
    \midrule[0.4pt]

    GPT-5.4
    & 64.6 & 37.7
    & 47.5 & 93.5
    & 57.2 & 85.4 & \textbf{77.5} & 29.8 & 92.3 & 48.7
    & 51.2 & 70.5 & 65.1
    \\

    Claude Sonnet 5
    & 54.1 & 8.1
    & 36.5 & 92.5
    & 23.5 & 55.8 & 51.3 & 14.3 & 89.7 & 45.4
    & 31.1 & 64.5 & 46.7
    \\

    Gemini 3.5 Flash
    & 65.9 & 40.2
    & 36.6 & \textbf{94.0}
    & 41.5 & 78.2 & 63.2 & 27.0 & 94.3 & 53.0
    & 53.1 & 65.3 & 59.5
    \\

    \midrule[0.7pt]
    \multicolumn{14}{c}{\textit{Open-Source Models}} \\
    \midrule[0.4pt]

    Qwen2.5-VL-7B
    & 37.3 & 36.8
    & 29.6 & 75.5
    & 19.7 & 34.2 & 23.0 & 1.6 & 92.3 & 36.5
    & 37.1 & 52.6 & 34.6
    \\

    Qwen3.5-4B
    & 36.9 & 30.8
    & 33.6 & 77.5
    & 22.5 & 38.6 & 48.4 & 8.2 & 93.3 & 34.3
    & 33.9 & 55.6 & 40.9
    \\

    Qwen3.5-9B
    & 42.2 & 34.6
    & 36.3 & 66.5
    & 19.7 & 47.2 & 47.2 & 10.4 & 93.7 & 39.7
    & 38.4 & 51.4 & 43.0
    \\

    VideoLLaMA3-7B
    & 34.2 & 34.1
    & 28.0 & 63.5
    & 18.5 & 46.8 & 0 & 0 & 91.7 & 37.3
    & 34.2 & 45.8 & 32.4
    \\

    InternVL3.5-8B
    & 41.1 & 34.1
    & 29.0 & 66.0
    & 16.7 & 38.6 & 13.1 & 0.5 & \textbf{95.0} & 42.3
    & 37.6 & 47.5 & 34.4
    \\

    Molmo2-8B
    & 41.6 & 36.8
    & 30.8 & 66.0
    & 18.8 & 32.8 & 21.5 & 0.1 & 91.0 & 36.9
    & 39.2 & 48.4 & 33.5
    \\

    \midrule[0.7pt]

    \rowcolor{gray!12}
    \textbf{BasketballSkills (Ours)}
    & \textbf{93.3} & \textbf{99.2}
    & \textbf{87.0} & 88.0
    & \textbf{69.8} & \textbf{91.8}
    & 47.8 & \textbf{45.9}
    & \textbf{95.0} & \textbf{83.7}
    & \textbf{96.3} & \textbf{87.5} & \textbf{72.3}
    \\

    \bottomrule[1.2pt]
\end{tabular}
\end{adjustbox}

\caption{Performance comparison across the ten BasketballBench tasks and
their modality-level averages. Scores are reported using the primary metric
of each task, and the best result in each column is highlighted in bold.
Overall scores are computed as macro-averages over the corresponding tasks.}
\vspace{-3pt}
\label{tab:main_results}
\end{table*}
\section{Experiments}
\begin{figure*}[t]
    \centering
    \includegraphics[width=\textwidth]{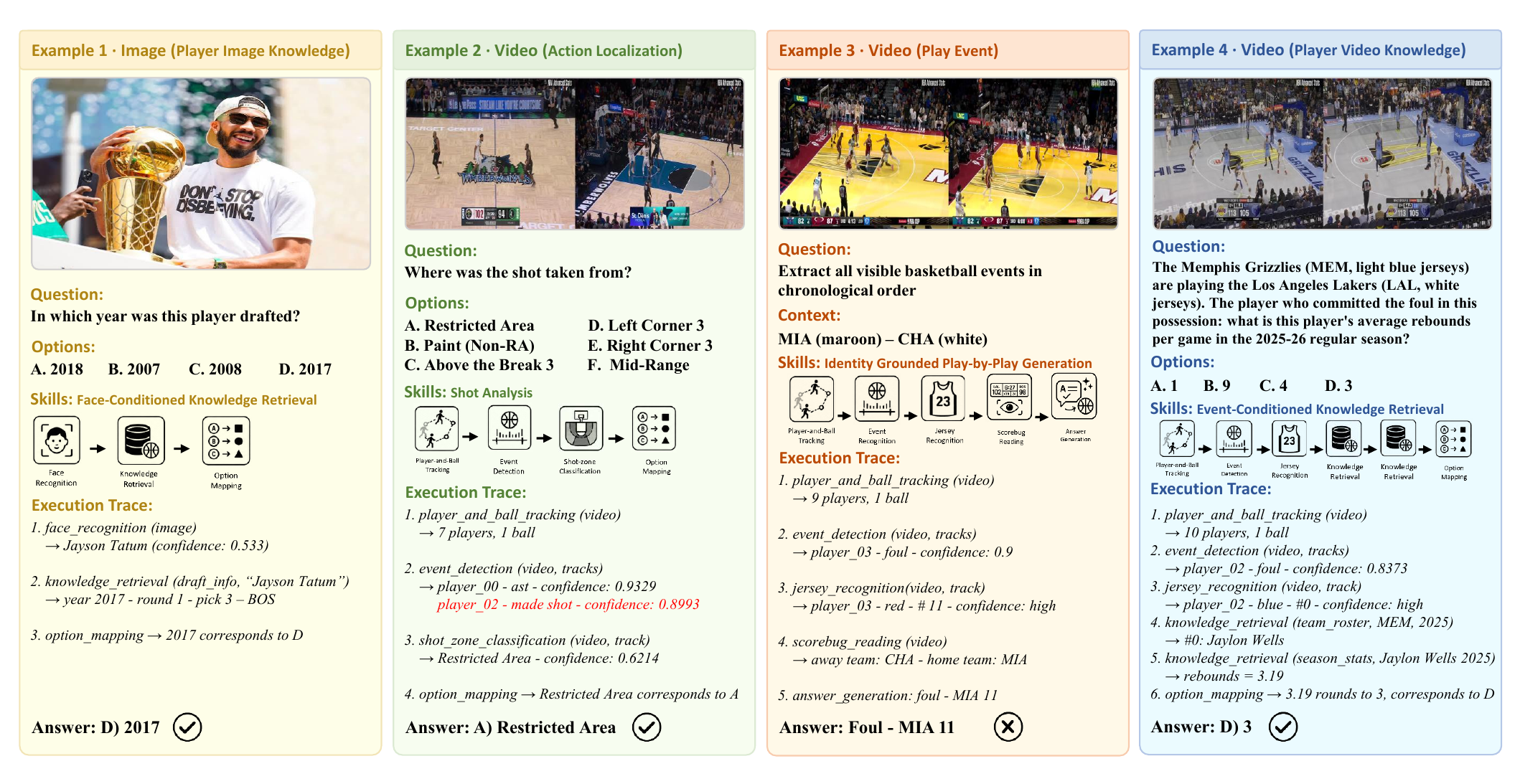}
    \vspace{-24pt}
    \caption{Representative BasketballSkills execution examples. Each example visualizes the selected skill, invoked tool sequence, intermediate observations, and final prediction. Both the successful and failed cases further illustrate how evidence is composed and how upstream errors may propagate to the answer.}
    \vspace{-6pt}
    \label{fig:basketballskills_trace}
\end{figure*}

We first describe the experimental setup in Section~\ref{subsec:experiments:settings}, then report the main results in Section~\ref{subsec:experiments:main}. Section~\ref{subsec:experiments:ablation}--\ref{subsec:experiments:qualitative} present ablations and representative execution traces.

\subsection{Experimental Setup}
\label{subsec:experiments:settings}
We present the experimental setup and main metrics for each benchmark task. Reproducibility details and additional metrics are in the \textbf{Supplementary Materials}.
\paragraph{Baselines and Inference Protocol.}
We compare BasketballSkills with three commercial MLLMs~(GPT-5.4~\cite{gpt54}, Claude Sonnet 5~\cite{claudesonnet5} and Gemini 3.5 Flash~\cite{gemini35flash}) and six open-source models~(Qwen3.5-4B, Qwen3.5-9B~\cite{qwen2026qwen35}, Qwen2.5-VL-7B~\cite{Qwen25VL}, VideoLLaMA3-7B~\cite{videollama3}, InternVL3.5-8B~\cite{internvl35}, and Molmo2-8B~\cite{molmo2}). 
All the videos are sampled at 2 fps, except that the Jersey Recognition tool uses uniformly sampled 8 crops of the same player. VideoLLaMA3-7B is evaluated on Q7 and Q8, but none of its outputs
satisfy the required task-specific schemas. Under our all-example
evaluation convention, these outputs receive zero credit rather than
being excluded from evaluation.

\paragraph{Output Formats and Metrics.}
Q1--Q3, Q6, Q9, and Q10 are four-way multiple-choice tasks, while Q5 uses six shot-zone choices; all are evaluated by accuracy.
Q4 returns structured scorebug fields and is evaluated by exact match.
Q7 contains two equally sized temporal-localization modes.
The video-timestamp mode predicts the elapsed time within the clip,
whereas the game-clock mode predicts the period and remaining game
time. Both modes are evaluated using all-example Acc@1s, treating
unparseable outputs as incorrect. The primary Q7 score is the
macro-average of video-timestamp and game-clock Acc@1s.
Q8 returns a chronological event sequence and is evaluated by full-event F1, where a match requires both the event type and participants to be correct.
Detailed evaluation protocols are provided in  the \textbf{Supplementary Materials}.

\subsection{Main Results}
\label{subsec:experiments:main}

\paragraph{General-purpose MLLMs.}
Table~\ref{tab:main_results} reveals strongly task-dependent performance among general-purpose MLLMs. \textbf{(1) Direct visual reading.} Their clearest strength lies in relatively direct tasks that primarily require OCR, such as Q4 and Q9. \textbf{(2) Player--event grounding.} They are much less effective at binding an event to the player involved. On Q6, a model must first identify the shooter associated with the event and then recover the player's jersey identity; every MLLM performs worse on this task than on Q9, which directly asks for a jersey number from player crops. The detailed Q8 results in Table~\ref{tab:q8_core_metrics} provide further evidence: event-type F1 is consistently much higher than full-event F1, for which the associated participants must also be correct. Moreover, when the event-relevant player is explicitly highlighted in Q5 and Q6, performance improves markedly across all evaluated MLLMs (Table~\ref{tab:target_player_grounding}), confirming that player--event grounding is a major bottleneck. \textbf{(3) Knowledge-intensive understanding.} General-purpose MLLMs perform considerably worse on tasks that require basketball knowledge or game-specific retrieval, particularly Q1--Q3 and Q10.

Overall, no general-purpose MLLM performs consistently across the benchmark: the best modality-level averages are only 53.1~$\%$ on TextQA, 70.5~$\%$ on ImageQA, and 65.1$\%$ on VideoQA. 
The performance gap is especially pronounced between relatively direct visual tasks, where scores reach 94.0~$\%$ on Q4 and 95.0~$\%$ on Q9, and more demanding tasks involving event composition or knowledge grounding, where the best results are only 29.8~$\%$ on Q8 and 53.0~$\%$ on Q10.

\paragraph{BasketballSkills.}
\textbf{(1) Overall performance.} BasketballSkills achieves the best result on eight of the ten tasks, including a tie on Q9. \textbf{(2) Knowledge-intensive tasks.} It substantially outperforms all general-purpose MLLMs on Q1--Q3. \textbf{(3) Compositional tasks.} Its advantages also extend to Q5, Q6, Q8, and Q10, which require different combinations of event recognition, player grounding, OCR, spatial understanding, and knowledge retrieval. These results show that BasketballSkills can select and compose complementary basketball-specific skills, enabling it to solve multi-stage tasks more effectively than a single general-purpose MLLM.
Notably, it surpasses the strongest general-purpose baseline by 59.0 percentage points on Q2, 39.5 percentage points on Q3, and 30.7 percentage points on Q10.

\begin{table}[t]
    \centering
    \renewcommand{\arraystretch}{1.08}
    \setlength{\tabcolsep}{3.0pt}
    \begin{adjustbox}{max width=\columnwidth}
    \footnotesize
    \begin{tabular}{@{}lrrrr@{}}
        \toprule[1.2pt]
        \textbf{Model}
        & \shortstack{\textbf{Full-event} \\ \textbf{F1}}
        & \shortstack{\textbf{Event-type} \\ \textbf{F1}}
        & \shortstack{\textbf{JSON} \\ \textbf{Validity}}
        & \shortstack{\textbf{Participant} \\ \textbf{Accuracy}}
        \\
        \midrule[0.8pt]
        \multicolumn{5}{c}{\textit{Commercial APIs}} \\
        \midrule[0.5pt]
        GPT-5.4          & 29.8 & 59.6 & 99.9 & 46.7 \\
        Claude-Sonnet-5        & 14.3 & 43.9 & 95.5 & 24.5 \\
        Gemini 3.5 Flash     & 27.0 & 51.8 & 97.8 & 49.4 \\
        \midrule[0.8pt]
        \multicolumn{5}{c}{\textit{Open-Source Models}} \\
        \midrule[0.5pt]
        Qwen2.5-VL-7B      & 1.6  & 27.5 & \textbf{100.0} & 5.1  \\
        Qwen3.5-4B      & 8.2  & 39.1 & \textbf{100.0} & 20.6 \\
        Qwen3.5-9B      & 10.4  & 45.8 & 99.9 & 22.2 \\
        VideoLLaMA3-7B  & 0.0 & 0.0 & 0.0 & 0.0 \\
        InternVL3.5-8B     & 0.5  & 11.0  & \textbf{100.0} & 2.7  \\
        Molmo2-8B          & 0.1  & 4.7  & \textbf{100.0} & 1.9  \\
        \midrule[0.8pt]
        \textbf{BasketballSkills}
                        & \textbf{45.9} & \textbf{60.9}
                        & \textbf{100.0} & \textbf{70.6} \\
        \bottomrule[1.2pt]
    \end{tabular}
    \end{adjustbox}
    \caption{Four complementary Q8 metrics. Full-event F1 requires both event type and participants to match, whereas event-type F1 ignores participant identity. 
    }
    \vspace{-12pt}
    \label{tab:q8_core_metrics}
\end{table}

\begin{table}[t]
    \centering
    \small
    \setlength{\tabcolsep}{3pt}
    \renewcommand{\arraystretch}{1.10}
    \resizebox{\columnwidth}{!}{%
    \begin{tabular}{@{}lccc@{\hspace{7pt}}ccc@{}}
        \toprule
        \multirow{2}{*}{\textbf{Model}}
        & \multicolumn{3}{c}{\textbf{Q5}}
        & \multicolumn{3}{c}{\textbf{Q6}} \\
        \cmidrule(lr){2-4}\cmidrule(lr){5-7}
        & Raw & Target & $\Delta$
        & Raw & Target & $\Delta$ \\
        \midrule
        GPT-5.4
        & 57.2 & \textbf{61.5} & +4.3
        & 85.4 & \textbf{97.0} & +11.6 \\

        Claude Sonnet 5
        & 23.5 & \textbf{33.7} & +10.2
        & 55.8 & \textbf{84.4} & +28.6 \\

        Gemini 3.5 Flash
        & 41.5 & \textbf{47.8} & +6.3
        & 78.2 & \textbf{94.6} & +16.4 \\

        Qwen3.5-9B
        & 19.7 & \textbf{36.3} & +16.7
        & 47.2 & \textbf{85.8} & +38.6 \\

        Molmo2-8B
        & 18.8 & \textbf{21.5} & +2.7
        & 32.8 & \textbf{45.0} & +12.2 \\

        InternVL3.5-8B
        & 16.7 & \textbf{17.3} & +0.7
        & 38.6 & \textbf{49.4} & +10.8 \\

        \midrule
        \textbf{Mean}
        & 29.6 & \textbf{36.4} & +6.8
        & 56.3 & \textbf{76.0} & +19.7 \\
        \bottomrule
    \end{tabular}%
    }
    \vspace{-3pt}
    \caption{Effect of target-player grounding on Q5 and Q6. \emph{Raw} uses the original video, whereas \emph{Target} highlights the ground-truth event-relevant player with a bounding box in every frame. $\Delta$ denotes the absolute improvement in percentage points, computed from unrounded accuracies.}
    \vspace{-15pt}
    \label{tab:target_player_grounding}
\end{table}

\subsection{Ablations}
\label{subsec:experiments:ablation}

\begin{table}[t]
    \centering
    \renewcommand{\arraystretch}{1.12}
    \setlength{\tabcolsep}{3.0pt}
    \small
    \begin{adjustbox}{max width=\columnwidth}
    \begin{tabular}{@{}lrrr@{\hspace{6pt}}rrr@{}}
        \toprule[1.2pt]
        \multirow{2}{*}{\textbf{Modality}}
        & \multicolumn{3}{c}{\textbf{Performance} ($\uparrow$)}
        & \multicolumn{3}{c}{\textbf{Tool Calls} ($\downarrow$)}
        \\
        \cmidrule(lr){2-4}
        \cmidrule(lr){5-7}
        & No Skills & Skills & Rel.\ $\Delta$
          & No Skills & Skills & Rel.\ $\Delta$
        \\
        \midrule[0.8pt]
        TextQA
            & 95.92 & \textbf{96.21} & +0.30\%
            & 2.32 & \textbf{2.26} & -2.36\% \\
        \rowcolor{gray!10}
        ImageQA
            & \textbf{92.25} & \textbf{92.25} & 0.00\%
            & 2.04 & \textbf{2.00} & -2.33\% \\
        VideoQA
            & \textbf{72.76} & 72.34 & -0.58\%
            & 4.36 & \textbf{3.63} & -16.74\% \\
        \bottomrule[1.2pt]
    \end{tabular}
    \end{adjustbox}
    \vspace{-3pt}
    \caption{Modality-level ablation of procedural skills in BasketballSkills.
    Values are macro-averaged over tasks within each modality. Relative change
    is computed from unrounded modality averages as
    $(\text{Skills}-\text{No Skills})/\text{No Skills}\times100\%$.}
    \vspace{-12pt}
    \label{tab:skill_ablation}
\end{table}

Table~\ref{tab:skill_ablation} demonstrates that procedural skills substantially streamline execution for complex multimodal queries.
The most pronounced gain appears on VideoQA, where the average number of tool calls decreases from 4.36 to 3.63 per query, corresponding to a $16.74\%$ reduction, while aggregate performance remains comparable at 72.34$\%$, indicating that procedural skills are most beneficial when coordination demands are high.
TextQA and ImageQA performance is likewise preserved, indicating that the skill-guided controller reduces unnecessary exploration without compromising its ability to solve the underlying tasks. 
These results highlight the value of procedural skills in organizing long, multi-tool reasoning workflows, with their efficiency advantage becoming particularly evident on video-based questions.

\subsection{Qualitative Results}
\label{subsec:experiments:qualitative}
Figure~\ref{fig:basketballskills_trace} presents the complete BasketballSkills workflow through several representative examples, including skill selection, tool execution, intermediate observations, and final answer generation. 
The third case illustrates a failure: BasketballSkills unnecessarily invokes Scorebug Reading to identify the two teams, reflecting an overly cautious tool-use strategy. 
The same example also produces an incorrect final answer because Track-Level Event Detection returns an erroneous result, illustrating how upstream tool failures can propagate through subsequent grounding and answer generation. Together, these cases reveal remaining challenges in both tool reliability and execution control.

\section{Conclusion}
We introduced \emph{BasketballBench}, a comprehensive multimodal benchmark comprising 7,980 questions across ten tasks spanning text, image, and video modalities. Our experiments reveal that current MLLMs remain limited in the fine-grained spatial and temporal perception of basketball games. Their performance degrades further on compositional questions that require multiple capabilities, such as jointly recognizing events and players, localizing actions in space and time, and incorporating basketball knowledge. To address these challenges, we developed \emph{BasketballSkills}, a hierarchical agent that organizes basketball-specific tools into reusable procedural skills. BasketballSkills outperforms general-purpose MLLMs, demonstrating the effectiveness of composing specialized perception and retrieval capabilities for comprehensive basketball understanding. Together, \emph{BasketballBench} and \emph{BasketballSkills} provide a foundation for advancing fine-grained, compositional multimodal understanding in basketball and other complex sports domains.

\bibliography{reference}
\clearpage
\onecolumn
\pagestyle{plain}
\raggedbottom

\makeatletter
\def\@float@HH#1[H]{\@float@Hx{#1}[!htbp]}
\setlength{\@fptop}{0pt}
\setlength{\@fpbot}{0pt plus 1fil}
\makeatother

\setlength{\textfloatsep}{12pt plus 2pt minus 4pt}
\setlength{\floatsep}{10pt plus 2pt minus 2pt}
\setlength{\intextsep}{10pt plus 2pt minus 2pt}
\renewcommand{\topfraction}{0.90}
\renewcommand{\bottomfraction}{0.80}
\renewcommand{\textfraction}{0.08}
\renewcommand{\floatpagefraction}{0.75}
\setcounter{topnumber}{4}
\setcounter{bottomnumber}{2}
\setcounter{totalnumber}{6}

\captionsetup[table]{
  width=\textwidth,
  font=small,
  labelfont=bf,
  justification=raggedright,
  singlelinecheck=false,
  skip=6pt
}

\let\addcontentsline\articleaddcontentsline

\begin{center}
    {\LARGE\bfseries Supplementary Material\par}
    \vspace{0.5em}
    {\Large Towards Comprehensive Basketball Understanding\par}
\end{center}

\vspace{1em}

\begingroup
\setcounter{tocdepth}{2}
\tableofcontents
\endgroup

\clearpage

\appendix
\makeatletter
\renewcommand{\subsubsection}{%
    \@startsection{subsubsection}{3}{\z@}%
    {-6pt plus -2pt minus -1pt}%
    {-1em}%
    {\normalfont\normalsize\bfseries}%
}
\makeatother

\setcounter{secnumdepth}{3}
\setcounter{tocdepth}{2}

\numberwithin{equation}{section}
\numberwithin{figure}{section}
\numberwithin{table}{section}

\appendix

\section{BasketballBench Construction and Specifications}
\label{app:basketballbench}

This section describes the data sources, structured knowledge base, common
construction protocol, and task-specific generation procedures of
BasketballBench.

\subsection{Data Sources}
\label{app:data_sources}

\BasketballBench\ integrates structured basketball records, official player
images, possession-level broadcasts with game metadata, play-by-play (PBP)
records, and game-specific uniform metadata. Instances are anchored in the
2025--2026 NBA season, with historical statistics and drafts retained where
needed. All sources are frozen before generation.

\paragraph{Structured Basketball Records.}
Sportradar's NBA API~\cite{nba_api} provides information on all 30 franchises,
including teams, coaches, players, season statistics, schedules, game and
period summaries, box scores, officials, injuries, drafts, and free agents.
The normalized SQLite database described in Section~\ref{app:database}
grounds the four knowledge-oriented tasks.

\paragraph{Player Images.}
NBA.com~\cite{nba_official} supplies one official headshot for each of 530 registered players, forming the face-recognition gallery. We additionally collect 400 evaluation photographs, each manually verified to depict the associated player and to be distinct from that player’s official gallery headshot.

\paragraph{Broadcast Videos and Associated Game Metadata.}
The video set comprises 2,501 possession-level broadcasts from the
BasketEvent test split~\cite{zhang2026basketeventunderstandingdidbasketball},
covering 33 games in 2025--2026. The \(1280\times720\), 25-fps clips last
3.48--18.96\,s (mean 9.45\,s; standard deviation 1.81\,s). Each clip is
linked to its game date, teams, and active rosters, including game-specific
jersey numbers. These metadata support PBP alignment and conversion of
identity annotations into observable participant labels; they are not model
inputs unless explicitly included in a task prompt.

\paragraph{Play-by-play Records.}
NBA.com~\cite{nba_official} provides 43,558 chronological PBP events from 266
games, including all 33 video-source games. Each record specifies the game
clock, action, participants, and, when applicable, shot outcome and location.
We use ten event types---made shot, missed shot, free throw, rebound,
turnover, foul, steal, assist, block, and jump ball---and six shot regions:
Restricted Area, In the Paint (Non-RA), Mid-Range, Above the Break 3, Left
Corner 3, and Right Corner 3. A clip may align with several consecutive PBP
events (Section~\ref{app:video_annotation}).

\paragraph{Game-specific Uniform Metadata.}
NBA LockerVision~\cite{nba_lockervision} provides reference images for the
uniform editions of all 30 teams and the edition assigned to each team in
each 2025--2026 game. GPT-5.4~\cite{gpt54} identifies dominant and secondary colors from
the static reference images; these descriptions are normalized to canonical
English labels and manually verified (Section~\ref{app:uniform_annotation}).
Joining the game assignment with team and roster metadata converts a PBP
identity into a visible color--number label (e.g., \code{white-23}); prompts
then map colors to team tricodes. LockerVision data and derived colors are
used only for offline construction and verification: they are neither stored
in the structured basketball knowledge base described in Section~\ref{app:database} nor exposed to tools or evaluated models. GPT-5.4 never
annotates or answers questions about benchmark broadcast clips.

\subsection{Structured Basketball Knowledge Base}
\label{app:database}

\paragraph{Overview.}
\code{NBA_DB} is a SQLite knowledge base comprising 18 normalized tables
derived from Sportradar records. It grounds the four knowledge-oriented
tasks while keeping generated questions, choices, visual annotations, and
sample mappings outside the database.

\subsubsection{Temporal Snapshot and Data Coverage}
\label{app:database-coverage}

The primary snapshot corresponds to the 2025--2026 NBA season and covers
games from the October 2025 preseason through Game~5 of the NBA Finals on
June 13, 2026. It contains 1,423 schedule records across preseason,
regular-season, in-season tournament, play-in, and postseason competition,
with detailed box scores for all 1,230 regular-season games. Historical
player-season statistics cover season years 2012--2025, and draft records
span 2003--2026. Table~\ref{tab:database-coverage} summarizes the
benchmark-relevant coverage.

\begin{table}[H]
    \centering
    \footnotesize

    \renewcommand{\arraystretch}{1.18}
    \setlength{\tabcolsep}{6pt}
    \arrayrulecolor{black!35}
    \setlength{\arrayrulewidth}{0.35pt}

    \begin{tabularx}{\linewidth}{
        @{}
        >{\raggedright\arraybackslash}m{0.205\linewidth}
        !{\color{black!35}\vrule width 0.35pt}
        >{\raggedright\arraybackslash}X
        !{\color{black!35}\vrule width 0.35pt}
        >{\raggedright\arraybackslash}m{0.185\linewidth}
        @{}
    }
        \hline
        \textbf{Data Category}
        & \textbf{Coverage}
        & \textbf{Scale} \\
        \hline

        Teams
        & All current NBA teams
        & 30 teams \\
        \hline

        Players
        & Player profiles, including a small number of
          box-score-only stub records
        & 633 players \\
        \hline

        Seasonal schedules
        & All 2025--2026 games across PRE, REG, IST, PIT, and PST
        & 1,423 games \\
        \hline

        Detailed game records
        & Team-, player-, and period-level box scores for all
          regular-season games
        & 1,230 games \\
        \hline

        Player seasons
        & Player--team affiliations by season and competition type,
          covering season years 2012--2025
        & 7,940 records \\
        \hline

        Player-season statistics
        & Season totals and per-game averages across the collected
          statistical fields
        & 15,880 rows \\
        \hline

        Draft records
        & NBA draft records from 2003 through 2026
        & 551 records \\
        \hline

        Officials
        & Referee profiles and regular-season game assignments
        & {82 officials, 3,714 assignments} \\
        \hline

    \end{tabularx}

    \caption{Coverage of the structured basketball knowledge base.}
    \label{tab:database-coverage}
\end{table}

\subsubsection{Logical Organization}
\label{app:database-organization}

The tables form four conceptual modules: team and player records (profiles,
rosters, coaches, jersey numbers, and drafts); season records (affiliations
and statistics); schedules (games and teams); and game details (summaries,
officials, box scores, and period statistics). Season totals and per-game
averages are separated so that templates can specify their statistical scope.

\subsubsection{Entity Relationships}
\label{app:database-relations}

Player profiles link to current rosters, drafts, and season-specific
affiliations, which in turn link to total and per-game statistics. Separating
current membership from historical affiliation prevents present-day team or
jersey data from being applied to another season. Schedules connect the home
and away teams to summaries, officials, box scores, and period records,
preserving the player, team, season, and game context of each fact.

\subsubsection{Use in Benchmark Construction}
\label{app:database-usage}

Templates specify the required entities, temporal constraints, relations,
and answer type. Basketball Knowledge QA retrieves profiles, rosters, drafts,
schedules, and season statistics; Match-Situation QA uses game and period
records. The two visual knowledge tasks first resolve a player and then
retrieve the requested fact. Questions, answers, and distractors are stored
separately, leaving \code{NBA_DB} as a reusable factual source.

\subsection{Benchmark Construction Protocol}
\label{app:benchmark-construction}

\subsubsection{General Generation Procedure}
\label{app:benchmark-generation}

Each task defines admissible source entities, answer relations, and an
unambiguous output schema. Before quota-based sampling, we remove missing or
unresolved fields, duplicate prompts, and cases admitting multiple answers.
Multiple-choice answers are derived from source records; type-compatible
distractors use other observed categorical values or nearby numerical values,
with no duplicate choices and balanced answer positions. Open-ended answers
are serialized from aligned annotations under a fixed schema.

Multimodal instances additionally require consistent links between visual
evidence and source records: image questions must consistently link the evaluation photograph, the official gallery headshot, the roster entry, and the database profile to the same player, while video questions require aligned PBP events, participant attributes, and game metadata.

\subsubsection{Video–Record Alignment and Temporal Annotation}
\label{app:video_annotation}

Each clip is first linked to its source game and the corresponding PBP interval, retaining the ordered events, participants, shot outcomes, and shot locations recorded in the official data. As an initial automatic alignment stage, we process all 4,244 possession clips in the BasketEvent test split before task-specific sampling. Each clip is sampled at 1 fps, and Qwen3.6-27B~\cite{qwen2026qwen36} reads the scorebug clock in every sampled frame, producing a per-second mapping between clip-relative video time and game-clock time. We retain only clips for which all sampled scorebugs are readable and the recovered clocks form a consistent countdown sequence; clips with clock stoppages, missing or unreadable scorebugs, or inconsistent OCR results are excluded.
Matching an event’s official PBP clock to this mapping provides an initial estimate of its clip-relative video timestamp. However, this automatically derived timestamp is not used as the final temporal ground truth. Official PBP timestamps may lag the visually observable occurrence of an event by one or two seconds, and the 1-fps OCR mapping introduces additional temporal quantization. We therefore manually reannotate all 790 target-event instances selected for Temporal Localization QA (Q7). For each instance, annotators inspect the clip and record both the clip-relative video timestamp of the visible event and the corresponding period and remaining game-clock value shown in the broadcast at that moment. These manually corrected annotations serve as the final reference labels for the video-timestamp and game-clock modes of Q7. The OCR-based mapping and official PBP timestamps are used only for candidate alignment and quality filtering.
Game-day rosters and uniform assignments convert source identities into team, jersey-number, and color attributes. Prompts expose the two team–color correspondences, allowing outputs to use observable team and jersey labels rather than hidden player identities.

\subsubsection{Uniform-Color Normalization}
\label{app:uniform_annotation}

Edition-level colors are mapped to a controlled English vocabulary and
joined with each game's home and away assignments. Tasks requiring color
grounding retain only games with unambiguous metadata for both teams.
Model-assisted descriptions are manually checked against the reference
images before use.

\subsection{Task Definitions and Generation}
\label{app:task-definitions}

The benchmark contains 7,980 instances: 2,400 text, 600 image, and 4,980
video questions. Below we summarize task-specific selection and answer
construction; shared rules follow Section~\ref{app:benchmark-construction}.
The examples transcribe released benchmark instances; line breaks and list
formatting are compacted for typesetting, while the options and output
constraints are preserved. Videos are identified by game and clip IDs because
the same assets are shared across video tasks.

\newenvironment{taskexample}{%
    \noindent\begin{minipage}{\linewidth}%
    \setlength{\FrameRule}{0.35pt}%
    \setlength{\FrameSep}{4pt}%
    \setlength{\OuterFrameSep}{4pt}%
    \begin{framed}%
    \scriptsize
    \setlength{\parindent}{0pt}%
    \setlength{\parskip}{1pt}%
}{%
    \end{framed}%
    \end{minipage}\par
}

\subsubsection{Q1: Basketball Knowledge QA}
\label{app:task-q1}

This text-only task contains 1,200 four-way questions from 47 manually
designed types spanning player profiles, team attributes, rosters and
coaches, drafts, 2025--2026 statistics, career aggregates, and schedules.
Candidates require valid entity links, and season questions distinguish
totals from per-game averages. Templates retrieve and type-normalize the
reference value; distractors use other observed categorical values or nearby
values of the same statistic. Examples are allocated across question types,
with answers balanced over A--D.

\begin{taskexample}
\textbf{Media.} None (text-only input).\par
\textbf{Prompt.} What is Deandre Ayton's career regular-season total rebounds?
A. 4453; B. 4742; C. 5101; D. 5113. Return only the letter of your choice (A,
B, C, or D), with no extra words.\par
\textbf{Ground truth.} \textbf{B} (4742).
\end{taskexample}

\subsubsection{Q2: Match-Situation QA}
\label{app:task-q2}

This task contains 1,200 four-way questions (40 per type) about specific
regular-season games. Its 30 types cover game context, final and period
scores, officials, attendance, lead changes and ties, box scores, leaders,
player statistics, and derived outcomes. Date and matchup must uniquely
identify a game with all required records. References join schedules with
summaries, officials, period scoring, and box scores; distractors are valid
same-domain names, venues, or nearby numerical values. Answer positions are
balanced.

\begin{taskexample}
\textbf{Media.} None (text-only input).\par
\textbf{Prompt.} On March 12, 2026, in the Indiana Pacers vs. Phoenix Suns
game, who was the top scorer of the game? A. Isaiah Jackson; B. Buddy Hield;
C. Jalen Suggs; D. Devin Booker. Return only the letter of your choice (A, B,
C, or D), with no extra words.\par
\textbf{Ground truth.} \textbf{D} (Devin Booker).
\end{taskexample}

\subsubsection{Q3: Player Image Knowledge QA}
\label{app:task-q3}

This task pairs a photo of an NBA player with one of 23 knowledge‑question types, covering player profiles, 2025–2026 statistics, draft history, and career aggregates. Each sample requires consistent roster, image, and database links, and replaces the player’s name with “this player” to force visual recognition before retrieval. All photos are human‑verified to ensure that they contain a clear frontal view of the player’s face and are distinct from the official headshots.
The 400 questions follow Q1's typed answer and distractor rules. Image assignment favors player diversity and unused evaluation photographs, with balanced answer positions.

\begin{taskexample}
\textbf{Media.} One player photograph.
\par
\textbf{Prompt.} How many NBA regular seasons has this player played in his
career? A. 10; B. 8; C. 2; D. 13. Return only the letter of your choice (A, B,
C, or D), with no extra words.\par
\textbf{Ground truth.} \textbf{C} (2).
\end{taskexample}

\subsubsection{Q4: Scorebug Reading QA}
\label{app:task-q4}

This task evaluates broadcast-graphic reading on 200 sampled frames. A
vision-language model initially parses both teams' scores, the period, and
game clock; incomplete parses or invalid team metadata are removed, and all
retained fields are manually verified. Half of the open-ended questions use
\code{AWAY-HOME, away_score-home_score}, and half use
\code{PERIOD, MM:SS}; references are serialized from the verified labels.

\begin{taskexample}
\textbf{Media.} One broadcast frame.\par
\textbf{Prompt.} Read the on-screen broadcast scorebug in this basketball
frame. The left side is the away team and the right side is the home team
(standard NBA layout). What is the current score? Report the away-team
tricode, home-team tricode, and both team scores. Output exactly one line in
the format \code{AWAY-HOME, away_score-home_score}; for example,
\code{LAL-MEM, 80-95}. Return only that line, with no extra words.\par
\textbf{Ground truth.} \code{DEN-POR, 65-62}.
\end{taskexample}

\subsubsection{Q5: Action Localization QA}
\label{app:task-q5}

This task retains possessions with exactly one made or missed shot whose PBP
location maps unambiguously to Restricted Area, In the Paint (Non-RA),
Mid-Range, Above the Break 3, Left Corner 3, or Right Corner 3. Each video
asks ``Where was the shot taken from?'' and presents the complete six-region
taxonomy. The 600 examples are class-balanced; multiple-shot possessions and
missing locations are excluded.

\begin{taskexample}
\textbf{Media.} One possession video.\par
\textbf{Prompt.} Where was the shot taken from? The six zones are: Restricted
Area (the semicircle under the rim); In The Paint (Non-RA) (the key excluding
that semicircle); Above the Break 3 (the top-center three-point arc); Left and
Right Corner 3 (the corresponding sideline--baseline intersections); and
Mid-Range (outside the paint but inside the arc). A. Restricted Area; B. In
The Paint (Non-RA); C. Above the Break 3; D. Left Corner 3; E. Right Corner 3;
F. Mid-Range. Return only the letter (A--F), with no extra words.\par
\textbf{Ground truth.} \textbf{D} (Left Corner 3).
\end{taskexample}

\subsubsection{Q6: Action Identification QA}
\label{app:task-q6}

This task asks which player took the possession's unique made or missed shot.
Eligible shooters require valid teams and jersey numbers, and both teams
require game-specific colors. Prompts provide the team--color mapping, and
choices use team tricodes plus jersey numbers. The ground truth shooter is paired with
one same-team and two opposing-team distractors, testing team discrimination
and jersey recognition. The 500 four-way questions have balanced answer
positions.

\begin{taskexample}
\textbf{Media.} One possession video.\par
\textbf{Prompt.} Who took the shot on this possession? The shooter is
identified by team tricode and jersey number (e.g., ``LAL 23''). Teams
(tricode: jersey color): LAL: white, MEM: light blue. A. LAL 2; B. LAL 1;
C. MEM 12; D. MEM 32. Return only the letter of your choice (A, B, C, or D),
with no extra words.\par
\textbf{Ground truth.} \textbf{C} (MEM 12).
\end{taskexample}

\subsubsection{Q7: Temporal Localization QA}
\label{app:task-q7}

This task covers made shots, missed shots, assists, blocks, rebounds, steals, turnovers, and fouls. A clip is eligible only when the queried event type occurs exactly once and the event can be localized unambiguously in the video. All selected target events are manually annotated at their visually observable occurrence. Each annotation records two synchronized temporal coordinates: the clip-relative video timestamp and the period plus remaining game-clock value displayed on the scorebug at that moment. Each of the 790 target-event instances is converted into two open-ended questions, one for each temporal mode, yielding 1,580 questions in total. The dataset is stratified by event class and question formulation.

\begin{taskexample}
\textbf{Media.} One possession video.\par
\textbf{Prompt (video-time variant).} Please locate the precise moment in the video when an assist
is made (a pass leading directly to a made basket), and output the
corresponding video timestamp for that exact instant. Please strictly follow
the output format example: \code{[timestamp_in_seconds]}, e.g., \code{[5.0]}.\par
\textbf{Ground truth.} \code{[7.0]}.
\end{taskexample}

\subsubsection{Q8: Play Event QA}
\label{app:task-q8}

This task represents an entire possession as ordered JSON over the ten-event
taxonomy. Every event requires resolvable team and jersey identities, and
the prompt supplies both team--color mappings. Assists, blocks, and steals
are emitted as distinct events. The 1,000 clips are sampled to preserve
coverage of rare assists, steals, and blocks before filling the remaining
quota.

\begin{taskexample}
\textbf{Media.} One possession video.\par
\textbf{Prompt.} You are given a basketball possession clip. Game context:
Team A is MIA in maroon; Team B is CHA in white. Extract all visible
basketball events in chronological order. Allowed event types are
\code{missed_shot}, \code{made_shot}, \code{rebound}, \code{assist},
\code{steal}, \code{turnover}, \code{block}, \code{free_throw},
\code{foul}, and \code{jump_ball}. Output only valid JSON with an
\code{events} list, where each item has \code{event_type} and
\code{participants}. Participants use \code{TRICODE-JERSEY_NUMBER}; the
tricode must be MIA, CHA, or \code{unclear}, and an unreadable number is
written as \code{unclear}. Every non-jump-ball event has one role-specific
participant---the shooter, rebounder, assisting passer, stealer, turnover
committer, blocker, free-throw shooter, or fouler---and \code{jump_ball} has
two. Repeat recurring event types in chronological order; if none occurs,
output \code{{"events": []}}. Do not add explanations, timestamps, confidence
scores, or text outside the JSON.\par
\textbf{Ground truth.}
\code{{"events": [{"event_type": "foul", "participants": ["MIA-24"]}]}}.
\end{taskexample}

\subsubsection{Q9: Jersey-Number Recognition QA}
\label{app:task-q9}

This task selects one sufficiently long, jersey-annotated player track per
possession and samples eight valid crops in temporal order. A four-way
question uses the annotated jersey as its answer and NBA-profile jersey
numbers as distractors. The 300 examples use distinct possession--player
pairs and balanced answer positions.

\begin{taskexample}
\textbf{Media.} Eight chronologically ordered player crops.\par
\textbf{Prompt.} These 8 images show crops of a single basketball player,
uniformly sampled across a possession clip. The images are ordered
chronologically, from the beginning to the end of the clip. What is the
jersey number worn by this player? A. 34; B. 10; C. 22; D. 35. Return only
the letter of your choice (A, B, C, or D), with no extra words.\par
\textbf{Ground truth.} \textbf{B} (10).
\end{taskexample}

\subsubsection{Q10: Player Video Knowledge QA}
\label{app:task-q10}

This task combines event-conditioned player recognition with knowledge
retrieval. A clip must contain a unique primary participant describable by an
event role (e.g., ``the player who made the shot''), linked to a database
profile. Ambiguous multi-team cases are
excluded where current-team membership is queried. Prompts provide the
team--color context and ask one of 35 four-way question types spanning
statistics, profiles, teams, drafts, and source-game box scores. Database
answers use same-type or nearby-value distractors. The 1,000 questions have
balanced answer positions.

\begin{taskexample}
\textbf{Media.} One possession video.\par
\textbf{Prompt.} In this clip, the New York Knicks (NYK, black jerseys) are
playing the Miami Heat (MIA, white jerseys). The player who missed the shot in
this possession: in which city is this player's team based? A. Denver;
B. Detroit; C. Cleveland; D. New York. Return only the letter of your choice
(A, B, C, or D), with no extra words.\par
\textbf{Ground truth.} \textbf{D} (New York).
\end{taskexample}

\FloatBarrier
\section{BasketballSkills Architecture and Implementation}
\label{app:basketballskills}

This section details \BasketballSkills\ using the paper-facing names of all
four skills and eight tools. Repository identifiers appear only in the
verbatim controller-visible prompts (Section~\ref{app:prompts}).

\subsection{Agent Architecture}
\label{app:agent-architecture}

A temperature-zero DeepSeek-V4-Flash controller~\cite{deepseekv4} interprets
the query, optionally loads a skill, invokes tools, and iterates over their
structured outputs. Skills specify which tool outputs are required and the
order in which tools should be called; they do not generate tool results.

Each run allows at most 12 controller turns and 10 tool calls. The lightweight
verifier mentioned in the main paper is a fixed Python program that checks
each function call before execution. It validates the tool name, required
arguments and their types, media and path availability, and duplicate calls;
a failed check is returned to the controller for revision. Media and artifact
paths are anonymized to hide source-game identifiers and resolved only at
execution. The trace records turns, verification, tools, skills, outputs, and
the final answer for reproducibility.

\subsection{Atomic Tool Library}
\label{app:tools}

The eight atomic tools expose typed, task-bounded interfaces. Four
trajectory-dependent video tools reuse frozen precomputed artifacts when
available and otherwise run online; caching affects latency, not semantics.

\subsubsection{Player-and-Ball Tracking}
\label{app:tool-tracking}

\paragraph{Implementation.}
Player-and-Ball Tracking combines SAM 3~\cite{carion2025sam3}, a fine-tuned
RF-DETR detector~\cite{robinson2026rfdetr}, and BoT-SORT association~\cite{aharon2022bot}. SAM 3
propagates player masks, while RF-DETR detects on-court players and the ball.
To remove referees, bench players, and staff, a SAM trajectory must overlap
an RF-DETR player detection by at least 0.3 IoU in at least half of its
observed frames; ball trajectories come directly from detector--tracker
output. The result stores normalized \((x,y,w,h)\) boxes, explicit missing
observations, stable track identifiers, and anomaly flags for unreliable
entity counts.

\paragraph{RF-DETR Training Data.}
BasketEvent~\cite{zhang2026basketeventunderstandingdidbasketball} provides
frame-level player and ball boxes plus track-level event labels; only the
boxes train the detector. Preserving the original partitions, we randomly
select 3,000/200/200 train/validation/test videos and uniformly sample one
frame per video. The resulting 3,400-image, two-class COCO dataset contains
28,266/2,069/1,856 boxes, respectively (32,191 total); the test set is held
out for final evaluation. 

\paragraph{Data Leakage Prevention.}
All \BasketballBench\ clips belong to the
BasketEvent test split and are therefore disjoint from every frame used for
detector training or validation.
\paragraph{RF-DETR Training and Evaluation.}
RF-DETR Medium is initialized from its official COCO checkpoint, given a
two-class head, and fully fine-tuned for 100 epochs on \(576\times576\)
inputs. Training uses batch size 16, four-step accumulation (effective 64),
a learning rate of \(10^{-4}\) for all non-encoder detector parameters
(including the newly initialized classification head), and
\(1.5\times10^{-4}\) for the DINOv2 encoder, with EMA decay 0.993 and seed 42.
One NVIDIA RTX 3090 requires approximately four hours.
The selected epoch-25 EMA checkpoint obtains 0.6409 validation mAP@50:95. On
the 200-image test set, it reaches 0.6504 mAP@50:95, 0.8709 mAP@50, 0.7303
mAP@75, and 0.7489 AR@100. Mapping the official COCO model's \emph{sports
ball} and \emph{person} classes yields 0.3882 mAP@50:95 under the same
evaluation. Table~\ref{tab:app-rfdetr-classwise-comparison} gives the
corresponding class-wise comparison.

\begin{table}[H]
    \centering
    \small
    \setlength{\tabcolsep}{8pt}
    \renewcommand{\arraystretch}{1.08}
    \begin{tabular*}{\textwidth}{@{\extracolsep{\fill}}lrr@{}}
        \toprule
        \textbf{Model} & \textbf{Basketball AP@50:95}
        & \textbf{Player AP@50:95} \\
        \midrule
        Original RF-DETR Medium  & 0.1734 & 0.6029 \\
        \textbf{Fine-tuned RF-DETR Medium} & \textbf{0.5122} & \textbf{0.7886} \\
        \midrule
        Improvement & +0.3388 & +0.1857 \\
        \bottomrule
    \end{tabular*}
    \caption{Class-wise RF-DETR Medium detection performance on the held-out
    test set. Both metric columns report AP@50:95. For the original COCO
    model, \emph{sports ball} and \emph{person} are mapped to basketball and
    player, respectively.}
    \label{tab:app-rfdetr-classwise-comparison}
\end{table}

\paragraph{Information Exposed to the Agent.}
The controller receives the trajectory- and run-artifact references, player
and ball counts, and anomaly flags, but no masks, service configuration, or
source-game identifier.

\subsubsection{Track-Level Event Detection}
\label{app:tool-event-detection}

\paragraph{Implementation.}
Track-Level Event Detection applies BasketEvent's
PlayNet~\cite{zhang2026basketeventunderstandingdidbasketball} to video and
participant trajectories, assigning track identifiers to made and missed
shots, free throws, fouls, turnovers, jump balls, rebounds, steals, blocks,
and assists. Duplicate predictions for one action are reduced to the
highest-confidence track and ordered by temporal midpoint. Internal
timestamps support ordering only; exact moments require Event Temporal
Localization.

\paragraph{Information Exposed to the Agent.}
Given the video and tracking artifact, the tool returns ordered track-level
records with canonical event type, class label, confidence, and top-three
alternatives.

\subsubsection{Shot-Zone Classification}
\label{app:tool-shot-zone}

\paragraph{Implementation.}
Shot-Zone Classification uses an author-implemented six-class head added to
PlayNet's pretrained PlayerEventModel. Given a made- or missed-shot track
identified by Track-Level Event Detection, it predicts one of the six
benchmark regions. It neither selects the shooter independently nor processes
isolated free throws.

\paragraph{Training Data.}
The six-way dataset joins BasketEvent videos and trajectories with official
PBP records. Each retained possession has one made or missed shot, an
unambiguous shooter trajectory that overlaps the ball in at least one frame,
and a valid court-area label. It contains 16,391 examples from 220 games:
13,966/206/2,219 examples from 185/2/33 games for train/validation/test. Class
counts are 5,386 Above the Break 3, 4,077 Restricted Area, 3,290 In the Paint
(Non-RA), 1,934 Mid-Range, 879 Left Corner 3, and 825 Right Corner 3,
motivating class-balanced training. 

\paragraph{Data Leakage Prevention.}
Because all benchmark clips come from
the BasketEvent test split, they are disjoint from every shot-zone training
and validation video.

\paragraph{Training and Evaluation.}
We add a six-class head to PlayNet's pretrained PlayerEventModel, reusing its
TimeSformer backbone, trajectory-conditioned features, and relation modules.
The event heads remain frozen; the new head and backbone are fine-tuned on 12
eight-frame clips per video at 4 fps and \(224\times224\). Training runs for
20 epochs on four NVIDIA H800 GPUs with distributed data parallelism, one
video per GPU, 16-step accumulation (effective batch 64), and AdamW. Head and
backbone learning rates are \(5\times10^{-5}\) and \(10^{-6}\), with weight
decay 0.05, gradient clipping 1.0, and seed 123. To address the training-set
imbalance, weighted cross-entropy uses 0.1 label smoothing and the class
weight
\[
    w_c=\operatorname{clip}\!\left(\frac{N}{6n_c},0.2,5.0\right),
\]
where \(n_c\) is the number of training examples in class \(c\) and
\(N=13{,}966\). Table~\ref{tab:app-shot-zone-class-weights} reports the exact
frequencies and weights used in training.

\begin{table}[H]
    \centering
    \small
    \setlength{\tabcolsep}{8pt}
    \renewcommand{\arraystretch}{1.08}
    \begin{tabular*}{\textwidth}{@{\extracolsep{\fill}}clrr@{}}
        \toprule
        \textbf{Class ID} & \textbf{Shot zone}
        & \textbf{Training examples} & \textbf{Weight} \\
        \midrule
        0 & Above the Break 3       & 4,592 & 0.5069 \\
        1 & Restricted Area         & 3,471 & 0.6706 \\
        2 & Left Corner 3           &   755 & 3.083  \\
        3 & Right Corner 3          &   695 & 3.3492 \\
        4 & Mid-Range               & 1,602 & 1.453  \\
        5 & In The Paint (Non-RA)   & 2,851 & 0.8164 \\
        \bottomrule
    \end{tabular*}
    \caption{Training-set class frequencies and loss weights for
    Shot-Zone Classification.}
    \label{tab:app-shot-zone-class-weights}
\end{table}

Training takes approximately 8.75 hours. The checkpoint selected by validation macro-F1 (epoch 13) attains
0.7885 validation top-1 and 0.7660 macro F1. On 2,219 test examples, it
obtains 0.7071/0.9148/0.9739 top-1/2/3 accuracy, 0.6813 macro precision,
0.7610 macro recall, 0.6935 macro F1, and 0.7100 weighted F1.

\paragraph{Information Exposed to the Agent.}
Given a video, tracking artifact, and shooter track, the tool returns the
canonical region, class label, track, confidence, and, when available, the
six-class distribution.

\subsubsection{Jersey Recognition}
\label{app:tool-jersey}

\paragraph{Implementation.}
In video mode, Jersey Recognition uniformly samples eight observations from
each requested track, crops the player with context, and applies a fixed
jersey-reading instruction to Qwen3.5-9B~\cite{qwen2026qwen35}. Multi-image
mode applies the same recognizer to supplied crops. Responses are normalized
to color, number, and confidence; invalid inputs are reported rather than
imputed. Returning color with number disambiguates opponents sharing a jersey
number and permits team mapping from question context.

\paragraph{Information Exposed to the Agent.}
For each track, the controller receives its identifier, color, number or
null, and confidence, without access to the internal instruction or raw VLM
output.

\subsubsection{Face Recognition}
\label{app:tool-face}

\paragraph{Implementation.}
Face Recognition uses \code{face_recognition}~\cite{geitgey2017facerecognition}
and an offline gallery of encoded official headshots. It returns the nearest
identity only at distance \(\leq0.6\), otherwise null; confidence is one minus
the best distance, clipped to \([0,1]\).

\paragraph{Information Exposed to the Agent.}
The controller receives only the matched full name and confidence; null
denotes insufficient identity evidence.

\subsubsection{Structured Basketball Knowledge Retrieval}
\label{app:tool-knowledge}

\paragraph{Implementation.}
Structured Basketball Knowledge Retrieval provides fixed query families over
the frozen \code{NBA_DB} (Section~\ref{app:database}), covering player
profiles, biographies, statistics, career highs, game logs, rosters, team
metadata, search, drafts, date-conditioned games, leaderboards, comparisons,
box scores, summaries, and officials. It returns structured records and
never arbitrary SQL. A year denotes the season start, competition defaults
to regular season, and omitted filters broaden rather than silently select a
result. Career aggregates combine all-season records; game details require a
game identifier or unambiguous date--team combination.

\paragraph{Information Exposed to the Agent.}
The controller sees the query family, typed parameters, records, and
\code{NBA_DB} source marker, but no credentials, schema, or answer mappings.

\subsubsection{Scorebug Reading}
\label{app:tool-scorebug}

\paragraph{Implementation.}
Scorebug Reading uses Qwen3.5-9B to read away/home tricodes, scores, period,
and remaining clock from an image or from the first and last video frames. A
fixed instruction encodes the dataset's left-away/right-home convention and
forbids guessing illegible fields; unreadable scorebugs return an error.

\paragraph{Information Exposed to the Agent.}
The controller selects image or video mode and receives typed fields plus a
formatted summary, not general access to the VLM or its instruction.

\subsubsection{Event Temporal Localization}
\label{app:tool-temporal}

\paragraph{Implementation.}
Event Temporal Localization uses Qwen3.5-9B for an already identified event;
it is not a detector. At 2 fps, video-time mode locates the event relative to
the first frame, while game-clock mode reads the period and remaining clock
at that moment. Neither time system is arithmetically converted into the
other, and a fixed prompt selects the output schema.

\paragraph{Information Exposed to the Agent.}
Given a video, known-event description, and mode, the controller receives a
timestamp or period--clock pair with confidence and a formatted answer.

\subsection{Composite Skill Library}
\label{app:skills}

The controller loads one of four skills when its recurring evidence pattern
matches the query.

\paragraph{Face-Conditioned Knowledge Retrieval.}
Face Recognition first identifies the player; Structured Basketball
Knowledge Retrieval then obtains the requested fact, preventing choices or
prior knowledge from replacing visual identity evidence.

\paragraph{Shot Analysis.}
Player-and-Ball Tracking and Track-Level Event Detection identify the shot
and shooter, followed by Shot-Zone Classification for location or Jersey
Recognition for identity using the supplied color--team mapping.

\paragraph{Identity-Grounded Play-by-Play Generation.}
Tracking and event detection produce an ordered sequence; one Jersey
Recognition call resolves all participant tracks, which are joined back to
events and serialized by team and jersey number.

\paragraph{Event-Conditioned Player Knowledge Retrieval.}
The skill grounds an event participant from jersey attributes and team
context before invoking Structured Basketball Knowledge Retrieval, thereby
preventing premature player queries.

\subsection{Complete Agent-Visible Prompts}
\label{app:prompts}

This section reproduces the text made available to the controller. Prompt
construction is dynamic: the fixed system template contains a rendered tool
catalog and, when skills are installed, a skill catalog. We therefore present
the exact template followed by the exact text used to fill those dynamic
components. Tool-internal VLM instructions are not included here because
they are not visible to the controller and cannot be edited or invoked as
general prompts.

\newenvironment{agentprompt}{%
    \setlength{\FrameRule}{0.5pt}%
    \setlength{\FrameSep}{4pt}%
    \begin{framed}%
    \footnotesize\ttfamily\raggedright
    \setlength{\parindent}{0pt}%
    \setlength{\parskip}{2pt}%
}{%
    \end{framed}%
}
\newcommand{\promptline}[1]{\noindent\detokenize{#1}\par}

\subsubsection{System-Prompt Template}
\label{app:system-prompt}

The following is the complete fixed system-prompt template. At run time,
\code{tool_list} and \code{skill_block} are replaced by the blocks shown in
Sections~\ref{app:tool-prompt} and~\ref{app:skill-prompts}, respectively.

\begin{framed}
\begin{Verbatim}
You are a basketball analysis assistant. Based on the user's question and available media (image/video/game context), call tools step by step to gather evidence, then answer the question in a single text message.

Available tools ({num_tools} total):
{tool_list}

Tool dependencies (important):
- detect_events / classify_shot_zone require tracks_path (produced by track_entities)
- recognize_jersey has two modes: video+tracks mode (requires tracks_path from track_entities) OR multi-image mode (media_paths of pre-cropped player images, no tracks needed)
- classify_shot_zone requires shooter_player_id (identify the shooter from detect_events' players list where event is Made Shot or Missed Shot)
{skill_block}
Workflow:
0. If <available_skills> are listed above and one matches this task, prefer calling the `skill` tool first and follow its instructions before calling other tools.
1. Analyze the question and decide which tools to call and in what order (guided by any loaded skill)
2. Call tools and observe the returned results
3. Based on the results, decide the next step: call more tools, or give the final answer
4. Finally, answer the question in a plain text message (no more tool calls)

Rules:
1. Tool use is optional: call tools only when they will add evidence you actually need; you may answer directly from already-gathered evidence or from your own knowledge when a question does not require tool-based evidence. Tool arguments must be real values you know - do not fabricate file paths
2. When referencing fields from a previous tool's result, use the actual value (do not use ${{...}} placeholders)
3. Tools may fail or return errors; if a tool fails, adjust parameters and retry or switch to another tool
4. Answer in the same language as the question (Chinese/English)
5. Answer only based on collected tool evidence (when tools were used) - do not fabricate facts not present in the evidence
6. If evidence is insufficient to answer, state so honestly and provide a reasonable fallback value matching the expected answer format
7. Call at most 10 tools; if the limit is reached without sufficient evidence, give the best answer with what you have
8. Do not call the same tool with the same arguments more than once - repeated identical calls waste time and return identical results. Reuse the result you already have
\end{Verbatim}
\end{framed}

The initial user message is constructed as follows, with all media paths
already anonymized:

\begin{agentprompt}
\promptline{Question: {question}}
\promptline{Media: {anonymized media description}}
\end{agentprompt}

\subsubsection{Agent-Visible Tool Prompt}
\label{app:tool-prompt}

The eight tools are rendered into \code{tool_list} in the following order.
This text is supplied together with the corresponding function-calling
schemas, which enforce the parameter types and required arguments described
in Section~\ref{app:tools}. Output schemas are not separately injected; the
controller learns the output semantics from the descriptions below and from
the returned observations.

\begin{agentprompt}
\promptline{- track_entities(video_path: string, backend: string) [required: ['video_path']]: Track basketball players and the ball through the video. Returns tracks_path (string, path to the trajectory JSON consumed by downstream tools), player_count (integer), ball_count (integer, 0 or 1), anomalies (array of strings, e.g. "too_many_players"), run_dir (string). The trajectory JSON keys players as "player_NN" and the ball as "ball".}

\promptline{- detect_events(video_path: string, tracks_path: string) [required: ['video_path', 'tracks_path']]: Detect basketball events for each tracked player. Returns players (array, one entry per non-blank player, ordered by event time from earliest to latest). Each entry has player_id (string, e.g. "player_01"), event (string: Made Shot/Missed Shot/Free Throw/Foul/Turnover/Jump Ball/Rebound/steal/block/ast), event_label (integer), confidence (float), top3 (array of [name, prob] pairs).}

\promptline{- classify_shot_zone(video_path: string, tracks_path: string, shooter_player_id: string) [required: ['video_path', 'tracks_path', 'shooter_player_id']]: Classify a shooter's shot attempt into one of six court zones. Returns shooter_player_id (string), shot_location (string: Restricted Area/In The Paint (Non-RA)/Above the Break 3/Left Corner 3/Right Corner 3/Mid-Range), shot_location_label (integer), confidence (float), probs (object mapping each zone name to its prob, optional).}

\promptline{- recognize_jersey(video_path: string, tracks_path: string, tracking_ids: array, media_paths: array) [required: []]: Recognize jersey color and number for selected players from a video+tracks pair or pre-cropped images. Returns players (array). Each entry has tracking_id (string, e.g. "player_00"; "image_0" in multi-image mode), jersey_color (string), jersey_number (integer or null), confidence (string like "high"/"low" from cache, or numeric from VLM). Note: jersey_number alone is NOT unique across the clip - opposing players may share the same number; the (jersey_color, jersey_number) pair is unique per player.}

\promptline{- recognize_face(image_path: string) [required: ['image_path']]: Identify an NBA player from a face in a single image by matching against precomputed face encodings. Returns name (string, the matched player's full name; null when no stored face clears the tolerance) and confidence (float in [0,1], computed as 1 - best_face_distance; 0.0 when no match).}

\promptline{- query_nba_database(query_type: string, player_name: string, player_names: array, year: integer, season_type: string, team_alias: string, keyword: string, college: string, position: string, date: string, game_id: string, stat: string, top_n: integer, limit: integer, offset: integer) [required: ['query_type']]: Query structured NBA records (stats, rosters, games, drafts) from the local SQLite database. Returns data (object whose shape varies by query_type) and source (string, "NBA_DB"); on failure an "error" field is returned instead. year is the season start year (2024 = 2024-25); season_type defaults to REG. Most optional parameters BROADEN the result when omitted: season_stats/game_log/team_roster without year return ALL seasons, team_info without team_alias lists ALL teams. To get career-spanning figures call season_stats once WITHOUT year and aggregate the returned seasons list yourself; there is no separate career-total query type, so do not re-query per season. Prefer one broad query over many narrow ones.}

\promptline{- read_scorebug(image_path: string, video_path: string) [required: []]: Read the broadcast scorebug from one image or a video's first+last frames. Image input returns away_team, home_team (tricode strings), away_score, home_score (integers), period (integer 1-4 or "OT"), game_clock (string like "05:42"), answer (formatted summary), input_type="image". Video input returns first_frame and last_frame (objects with the above fields), answer (formatted "First: ... | Last: ..."), input_type="video". When the input has no readable scorebug, returns an error ("no scorebug visible in the input") instead of null fields.}

\promptline{- ground_event_time(video_path: string, event_description: string, time_mode: string) [required: ['video_path', 'event_description']]: It only localizes an event you already know happened. It CANNOT detect events from the video. time_mode="video_time" returns timestamp_seconds (float, seconds from video start), confidence (string), answer (string like "[5.0]"), time_mode="video_time". time_mode="game_clock" returns period (integer 1-4 or "OT"), game_clock (string read from the scorebug at that instant), confidence (string), answer (string like "[3 - 05:42]"), time_mode="game_clock".}
\end{agentprompt}

\subsubsection{Agent-Visible Skill Prompts}
\label{app:skill-prompts}

The skill catalog exposes the following repository names and descriptions.
These map respectively to the paper-facing skills Shot Analysis,
Event-Conditioned Player Knowledge Retrieval, Face-Conditioned Knowledge
Retrieval, and Identity-Grounded Play-by-Play Generation.

\begin{agentprompt}
\promptline{analyze-shot-attempt: Analyze a basketball shot attempt in video to determine its court zone or identify the shooter by team and jersey number. Use for shot-zone classification and shooter identification questions that require event detection followed by shot-specific analysis.}

\promptline{answer-event-conditioned-player-knowledge: Answer an NBA knowledge question about the player who performs a specified event in a basketball video. Use when solving the question requires event-role detection, jersey-based player resolution, and an NBA database lookup.}

\promptline{answer-face-conditioned-player-question: Identify a basketball player from a face image and answer an NBA knowledge question about that player. Use only when the question requires both face recognition and an NBA database lookup; do not use for database-only, jersey-only, or video-event questions.}

\promptline{generate-identity-grounded-play-by-play: Generate identity-grounded basketball play-by-play from a raw possession video by orchestrating entity tracking, event-role detection, jersey recognition, and context-provided team mapping. Use when the output requires chronological basketball events whose participants are represented by on-court identities such as team and jersey number; do not use for event labels, jersey recognition, or temporal localization alone.}
\end{agentprompt}

When a skill is selected, its body is returned inside a
\code{skill_content} wrapper. The following four blocks reproduce the full
provided \code{SKILL.md} texts.

\paragraph{Shot Analysis prompt.}\mbox{}\par
\begin{framed}
\begin{Verbatim}
---
name: analyze-shot-attempt
description: Analyze a basketball shot attempt in video to determine its court zone or identify the shooter by team and jersey number. Use for shot-zone classification and shooter identification questions that require event detection followed by shot-specific analysis.
---

# Analyze a Shot Attempt

## Workflow

1. Call `track_entities` on the video.
2. Pass the tracks to `detect_events`.
3. Select the relevant `Made Shot` or `Missed Shot` event and use its shooter tracking ID.
4. Follow the branch required by the question:
   - For a shot-zone question, call `classify_shot_zone` with the shooter tracking ID.
   - For a shooter-identification question, call `recognize_jersey` for the shooter tracking ID.
5. For shooter identification, map the recognized jersey color to the team mapping explicitly provided in the question, then combine the team identifier with the jersey number.
6. Return exactly the requested label or option.

## Constraints

- Obtain the shooter role only from `detect_events`.
- Do not select a player by visual prominence or proximity to the basket.
- Treat jersey identity as the pair `(jersey_color, jersey_number)`.
- Do not infer team identity from general NBA knowledge.
- Do not call `query_nba_database` unless the question explicitly requests external player knowledge.
\end{Verbatim}
\end{framed}

\paragraph{Event-Conditioned Player Knowledge Retrieval prompt.}\mbox{}\par
\begin{framed}
\begin{Verbatim}
---
name: answer-event-conditioned-player-knowledge
description: Answer an NBA knowledge question about the player who performs a specified event in a basketball video. Use when solving the question requires event-role detection, jersey-based player resolution, and an NBA database lookup.
---

# Answer an Event-Conditioned Player Knowledge Question

## Workflow

1. Call `track_entities` on the video.
2. Pass the tracks to `detect_events`.
3. Select the requested event and participant role, such as shooter, rebounder, fouler, or free-throw shooter.
4. Call `recognize_jersey` for that participant's tracking ID.
5. Map the jersey color to the team using only information supplied in the question.
6. Query the team roster with `query_nba_database`.
7. Resolve the player using the pair `(team, jersey_number)`.
8. Query `query_nba_database` for the fact requested about the resolved player.
9. Return exactly the requested answer format.

## Constraints

- Obtain the event role only from `detect_events`.
- Do not treat a jersey number alone as a unique player identity.
- Do not infer identity from answer choices, reputation, appearance, or basketball knowledge.
- Do not query a player fact before resolving the player from team and jersey number.
- Do not use `recognize_face` unless the question explicitly provides a suitable face image.
- Stop rather than guess if the event participant, team, jersey number, or roster identity cannot be resolved.
\end{Verbatim}
\end{framed}

\clearpage
\paragraph{Face-Conditioned Knowledge Retrieval prompt.}\mbox{}\par
\begingroup
\setlength{\OuterFrameSep}{0pt}
\begin{framed}
\begin{Verbatim}
---
name: answer-face-conditioned-player-question
description: Identify a basketball player from a face image and answer an NBA knowledge question about that player. Use only when the question requires both face recognition and an NBA database lookup; do not use for database-only, jersey-only, or video-event questions.
---

# Answer a Face-Conditioned Player Question

## Workflow

1. Call `recognize_face` on the provided face image.
2. Treat the returned player name as the only valid identity evidence.
3. If no player is recognized, stop rather than infer identity from the question or answer choices.
4. Call `query_nba_database` with the recognized name and the query type appropriate for the requested fact.
5. Derive the answer only from the database result.
6. Return exactly the format requested by the user.

## Constraints

- Do not identify the player visually without `recognize_face`.
- Do not infer identity from answer choices.
- Do not substitute basketball knowledge for missing tool evidence.
- Do not call tracking, event, jersey, or shot-zone tools.
\end{Verbatim}
\end{framed}
\endgroup

\paragraph{Identity-Grounded Play-by-Play Generation prompt.}\mbox{}\par
\begingroup
\setlength{\OuterFrameSep}{0pt}
\begin{framed}
\begin{Verbatim}
---
name: generate-identity-grounded-play-by-play
description: Generate identity-grounded basketball play-by-play from a raw possession video by orchestrating entity tracking, event-role detection, jersey recognition, and context-provided team mapping. Use when the output requires chronological basketball events whose participants are represented by on-court identities such as team and jersey number; do not use for event labels, jersey recognition, or temporal localization alone.
---

# Generate Identity-Grounded Play-by-Play

Transform a raw possession video into chronological play-by-play whose participants have on-court basketball identities.

## Workflow

1. Call `track_entities` on the raw video.
2. Pass the tracks to `detect_events`.
3. Treat the returned event types, roles, track IDs, and ordering as authoritative.
4. Collect the unique, non-null track IDs assigned to event roles.
5. Call `recognize_jersey` once for those track IDs.
6. Join each jersey result to its event role strictly by track ID.
7. Map each jersey color to a team using only the mapping supplied in the question.
8. Represent each participant using the required basketball identity, normally `(team, jersey_number)`.
9. Serialize the identity-grounded events exactly according to the requested play-by-play schema.

## Evidence Rules

- Obtain event types, roles, and ordering only from `detect_events`.
- Obtain jersey colors and numbers only from `recognize_jersey`.
- Treat `(jersey_color, jersey_number)` as the visual identity; a number alone is not unique.
- Use question-provided context as the only source of the color-to-team mapping.
- Reuse one recognized identity when the same track ID appears in multiple events.

## Boundaries

- Do not independently detect, add, remove, merge, or reorder events.
- Do not infer a team from NBA knowledge or uniform familiarity.
- Do not infer a participant from answer choices.
- Do not resolve real player names unless the requested output explicitly requires them.
- Do not include track IDs, confidence scores, or intermediate evidence in the final play-by-play unless requested.
- Preserve uncertainty when a required participant identity cannot be grounded.
\end{Verbatim}
\end{framed}
\endgroup

\FloatBarrier
\section{Additional Evaluation Results}
\label{app:experiments}

We define the non-multiple-choice metrics, make the treatment of unparsable
outputs explicit, and report fine-grained results. Multiple-choice tasks use
option accuracy; other scores are percentages and temporal errors are seconds
unless noted otherwise.

\subsection{Evaluation Metrics}
\label{app:metrics}
This subsection defines how parsing failures are treated for Scorebug Reading
(Q4), Temporal Localization QA (Q7), and Play Event QA (Q8), the three
non-multiple-choice tasks in BasketballBench.

\paragraph{Parsed versus All-example Evaluation.}
For an accuracy-based task with \(N\) examples, let \(N_{\mathrm{parsed}}\)
be the number of outputs that can be mapped to the task schema and let
\(N_{\mathrm{correct}}\) be the number of correct parsed outputs. We distinguish
\begin{equation}
    \mathrm{ParseRate}=\frac{N_{\mathrm{parsed}}}{N},\qquad
    \mathrm{ParsedAcc}=\frac{N_{\mathrm{correct}}}{N_{\mathrm{parsed}}},\qquad
    \mathrm{Acc}=\frac{N_{\mathrm{correct}}}{N}.
    \label{eq:parse-accuracy-definitions}
\end{equation}
Thus, \(\mathrm{Acc}=\mathrm{ParseRate}\times\mathrm{ParsedAcc}\), and
unparsable, missing, or failed responses receive zero credit in
\(\mathrm{Acc}\).

\subsubsection{Scorebug Reading (Q4)}
\label{app:metric-scorebug}

Scorebug Reading has equal score and clock subsets. A score prediction is
parsed as
\((t_a,t_h,s_a,s_h)\), containing the away and home team tricodes and their
scores. Team accuracy and score accuracy are slot-level measures:
\begin{equation}
    \mathrm{Acc}_{\mathrm{team}}
    = \frac{\text{number of correct team slots}}{2N_s},\qquad
    \mathrm{Acc}_{\mathrm{score}}
    = \frac{\text{number of correct score slots}}{2N_s},
\end{equation}
where \(N_s\) is the number of score prompts. Tricodes are uppercased and
scores converted to integers before comparison.

A clock prediction is correct only when its period and normalized
\(\mathrm{MM{:}SS}\) both match. The primary Q4 metric, Full Exact Match,
likewise requires all four score fields or both clock fields to match:
\begin{equation}
\mathrm{Acc}_{\mathrm{full}}
=\frac{
\sum_{i\in\mathcal{S}}\mathbf{1}[\hat{\mathbf{t}}_i=\mathbf{t}_i
\land\hat{\mathbf{s}}_i=\mathbf{s}_i]
+\sum_{i\in\mathcal{C}}\mathbf{1}[\hat p_i=p_i\land\hat c_i=c_i]
}{|\mathcal{S}|+|\mathcal{C}|}.
\end{equation}
This all-example Full Exact Match is the Q4 score in Table 2 of the main paper: an
unparseable or missing response is counted as incorrect. The same convention
is used for the three field-level scores in
Table~\ref{tab:app-scorebug-details}: every field in an unparseable output is
evaluated as false.

\subsubsection{Temporal Localization QA (Q7)}
\label{app:metric-action-spotting}

Video-timestamp mode predicts elapsed clip seconds; game-clock mode predicts
the period and remaining \(\mathrm{MM{:}SS}\). Acc@1s requires absolute error
of at most one second and, for game-clock predictions, a matching period.
Regulation periods match exactly, while all overtime labels are equivalent:
\begin{equation}
    \mathrm{ParsedAcc@1s}_{m}
    =\frac{1}{N^{\mathrm{parsed}}_m}
      \sum_i \mathbf{1}[d_i\leq 1],
    \qquad m\in\{\mathrm{video},\mathrm{clock}\},
\end{equation}
with the period constraint included in the indicator for game-clock mode.
The corresponding all-example score replaces the denominator by \(N_m\),
so unparsed outputs receive zero credit. 

The primary Q7 metric reported in Table~2 is the macro-average
of the all-example Acc@1s scores for the video-timestamp and
game-clock modes:
\begin{equation}
\mathrm{Q7}
=
\frac{1}{2}
\left(
\mathrm{Acc@1s}_{\mathrm{video}}
+
\mathrm{Acc@1s}_{\mathrm{clock}}
\right).
\end{equation}
Each mode contains 790 examples, so this macro-average is equivalent
to accuracy over all 1,580 Q7 questions. An output that remains
missing or unparseable after normalization receives zero credit.

Table~\ref{tab:app-action-details} additionally reports Parsed Acc@1s,
Parse Rate, and MAE. Video-timestamp MAE averages all parsed timestamps. Game-clock MAE averages
remaining-time error only when the period matches; mismatched periods remain
incorrect for Acc@1s but are excluded from MAE to avoid arbitrary
cross-period conversion.

\paragraph{Period-Alias Normalization.}
The requested game-clock format uses a numeric regulation-period identifier,
for example, \texttt{[1 - 06:53]}. We observe a recurrent parsing failure in
which a model instead emits the semantically equivalent ordinal form
\texttt{[1ST - 06:53]}. The clock value and period are both present, but a
minor label-format mismatch can make a substantial fraction of predictions
unparseable. For the fine-grained evaluation, we therefore normalize
\texttt{1ST}/\texttt{2ND}/\texttt{3RD}/\texttt{4TH} to
\texttt{1}/\texttt{2}/\texttt{3}/\texttt{4}. This rule is deterministic,
applied identically to every model, defined solely from the output schema, and
independent of the reference timestamp. It changes neither the predicted
period nor the clock value. Values that change under this repair are marked
with \(^{*}\) in Table~\ref{tab:app-action-details}.

\subsubsection{Play Event QA (Q8)}
\label{app:metric-event-extraction}

\paragraph{LCS-based event F1.}
Each Q8 output is a chronologically ordered list of event records,
where each event contains an \texttt{event\_type} and a
\texttt{participant}. We evaluate the predicted and reference
sequences under two matching rules:

\begin{itemize}
    \item \textbf{Type:} two events match if their event types are
    identical.
    \item \textbf{Type+Participant:} two events match only if both
    their event types and participants are identical.
\end{itemize}

For each rule, we use the longest common subsequence (LCS) to count
correctly ordered matches. LCS gives partial credit for correctly
predicted events while penalizing missing, additional, and incorrectly
ordered events. For example, if the reference sequence is
\([\text{shot},\text{rebound},\text{foul}]\) and the prediction is
\([\text{shot},\text{foul}]\), the LCS contains two matched events.

We sum the LCS match counts over all evaluation clips before computing
micro-averaged precision, recall, and F1:
\[
P_r=\frac{M_r}{N_{\mathrm{pred}}},
\qquad
R_r=\frac{M_r}{N_{\mathrm{ref}}},
\qquad
F1_r=\frac{2P_rR_r}{P_r+R_r},
\]
where \(M_r\) is the total number of LCS matches under rule
\(r\in\{\mathrm{Type},\mathrm{Type+Participant}\}\), and
\(N_{\mathrm{pred}}\) and \(N_{\mathrm{ref}}\) are the total numbers
of predicted and reference events. Metrics with a zero denominator
are defined as zero.

The primary Q8 metric is \(\mathrm{F1}_{\mathrm{Type+Participant}}\),
which evaluates event extraction and participant grounding jointly.
We additionally report \(\mathrm{F1}_{\mathrm{Type}}\) to isolate
event-type recognition.

\paragraph{Participant Accuracy.}
Participant Accuracy (PA) measures participant identification after
the event type has already been matched. Using the event pairs
selected by the Type-only LCS alignment, PA is the fraction for which
the predicted participant also matches the reference participant:
\[
\mathrm{PA}
=
\frac{
\#\text{type-aligned pairs with the correct participant}
}{
\#\text{type-aligned event pairs}
}.
\]
PA does not directly penalize missing, additional, or incorrectly
typed events and is therefore reported only as a diagnostic metric.

\paragraph{Conservative JSON Recovery.}
Some models produce recognizable event records in malformed JSON.
We apply a deterministic, model-agnostic recovery procedure that
extracts embedded JSON, removes trailing commas, completes only
unambiguous closing delimiters, and wraps a root event list in the
required \texttt{events} object. For truncated responses, only fully
closed event records are retained. Recovery never changes predicted
event types, participants, or event order and never consults the
reference answer.

The Q8 metrics are computed after this recovery step. An output that
remains invalid is treated as an empty predicted sequence. Models for
which recovery changes at least one output are marked with
\(^{*}\) in Table~\ref{tab:app-event-details}. The table additionally
reports Type-only metrics, Participant Accuracy, and the post-recovery
JSON validity rate.

\subsection{Fine-Grained Results}
\label{app:fine-grained-results}

This subsection presents a more fine-grained breakdown of the experimental results reported in Table 2 of the main paper.

\subsubsection{Scorebug Reading (Q4)}
\label{app:result-scorebug}

\begin{table}[H]
    \centering
    \small
    \renewcommand{\arraystretch}{1.06}

    \begin{tabular}{
        @{}l
        @{\hspace{1.5cm}}r
        @{\hspace{1.2cm}}r
        @{\hspace{1.2cm}}r
        @{}
    }
        \toprule
        \textbf{Model}
        & \textbf{Team Acc.}
        & \textbf{Score Acc.}
        & \textbf{Clock Acc.}
        \\
        \midrule

        GPT-5.4
        & 95.5 & 99.0 & 97.0
        \\

        Claude Sonnet 5
        & 94.5 & 99.0 & 97.0
        \\

        Gemini 3.5 Flash
        & 95.5 & 98.0 & 97.0
        \\

        \midrule

        Qwen2.5-VL-7B
        & 72.0 & 96.0 & 95.0
        \\

        Qwen3.5-4B
        & 76.5 & 98.0 & 91.0
        \\

        Qwen3.5-9B
        & 85.5 & 96.0 & 59.0
        \\

        VideoLLaMA3-7B
        & 42.0 & 91.0 & 95.0
        \\

        InternVL3.5-8B
        & 47.0 & 93.0 & 96.0
        \\

        Molmo2-8B
        & 55.5 & 90.5 & 92.0
        \\

        \midrule

        \textbf{BasketballSkills}
        & 88.5 & 98.0 & 97.0
        \\

        \bottomrule
    \end{tabular}

    \caption{
        Detailed Scorebug Reading results (\%).
        Team and Score are all-example slot accuracies, and Clock
        requires an exact match of both the period and normalized
        clock value. An unparseable or missing output is evaluated
        as incorrect for every corresponding field.
    }
    \label{tab:app-scorebug-details}
\end{table}

Table~\ref{tab:app-scorebug-details} reports the team identification, score recognition, and game-clock recognition performance of MLLMs and BasketballSkills on the Scorebug Reading task. Team identification exhibits the lowest accuracy, primarily because scorebugs often represent teams using logos, which MLLMs struggle to associate with the correct teams. Parse Rates are generally high for Scorebug Reading, so Parsed
Accuracy and all-example Accuracy are close for most models. We therefore
report only the more conservative all-example field accuracies, for which an
unparseable output is directly evaluated as incorrect.

\subsubsection{Action Localization QA (Q5)}
\label{app:result-shot-zone}

\begin{table}[H]
    \centering
    \footnotesize
    \setlength{\tabcolsep}{4.0pt}
    \renewcommand{\arraystretch}{1.08}
    \begin{tabular*}{\textwidth}{@{\extracolsep{\fill}}lrrrrrrr@{}}
        \toprule
        \textbf{Ground truth} & \textbf{RA} & \textbf{Paint}
        & \textbf{ATB3} & \textbf{LC3} & \textbf{RC3}
        & \textbf{Mid} & \textbf{U} \\
        \midrule
        Restricted Area & \textbf{74.0} & 6.0 & 1.5 & 1.0 & 0.0 & 9.0 & 8.5 \\
        Paint (Non-RA)  & 30.5 & \textbf{44.0} & 0.0 & 0.0 & 0.0 & 21.0 & 4.5 \\
        Above Break 3   & 1.0 & 0.0 & \textbf{64.0} & 15.0 & 4.0 & 16.0 & 0.0 \\
        Left Corner 3   & 2.5 & 2.0 & 0.0 & \textbf{90.0} & 0.0 & 4.0 & 1.5 \\
        Right Corner 3  & 3.0 & 0.0 & 1.0 & 2.0 & \textbf{83.0} & 8.0 & 3.0 \\
        Mid-Range       & 8.5 & 16.0 & 4.0 & 3.0 & 3.0 & \textbf{64.0} & 1.5 \\
        \bottomrule
    \end{tabular*}

    \vspace{2pt}
    \parbox{\textwidth}{\footnotesize RA: Restricted Area; Paint:
    In The Paint (Non-RA); ATB3: Above the Break 3; LC3/RC3: Left/Right
    Corner 3; Mid: Mid-Range.}
    \caption{Element-wise average of the row-normalized confusion matrices from two independent BasketballSkills runs on Action Localization QA. Rows are ground-truth
    zones and columns are predictions; U denotes an unparsed response.}\label{tab:app-zone-confusion}
\end{table}

As shown in Table~\ref{tab:app-zone-confusion}, Corner Threes are most distinctive (90.0\% left; 83.0\% right), whereas In
The Paint (Non-RA) is hardest (44.0\%), chiefly confused with adjacent
Restricted Area and Mid-Range regions.

\subsubsection{Temporal Localization QA (Q7)}   
\label{app:result-action-spotting}

\begin{table}[!htbp]
    \centering
    \footnotesize
    \setlength{\tabcolsep}{2.5pt}
    \renewcommand{\arraystretch}{1.08}

    \begin{tabular*}{\textwidth}{
        @{\extracolsep{\fill}}
        l
        rrrr
        rrrr
        @{}
    }
        \toprule

        \multirow{2}{*}{\textbf{Model}}
        & \multicolumn{4}{c}{\textbf{Video Timestamp}}
        & \multicolumn{4}{c}{\textbf{Game Clock}}
        \\

        \cmidrule(lr){2-5}
        \cmidrule(lr){6-9}

        & \textbf{P-Acc@1s}
        & \textbf{Acc@1s}
        & \textbf{Parse}
        & \textbf{MAE}
        & \textbf{P-Acc@1s}
        & \textbf{Acc@1s}
        & \textbf{Parse}
        & \textbf{MAE}
        \\

        \midrule

        GPT-5.4
        & 69.2 & 69.2 & 100.0 & 1.67
        & 86.2 & 85.7 & 99.4 & 1.52
        \\

        Claude Sonnet 5
        & 36.8 & 36.8 & 100.0 & 2.54
        & 65.8\(^{*}\) & 65.7\(^{*}\)
        & 99.9\(^{*}\) & 11.44\(^{*}\)
        \\

        Gemini 3.5 Flash
        & 57.6 & 55.9 & 97.2 & 1.30
        & 74.4 & 70.4 & 94.6 & 2.92
        \\

        \midrule

        Qwen2.5-VL-7B
        & 1.1 & 1.1 & 100.0 & 5.96
        & 45.0\(^{*}\) & 44.9\(^{*}\)
        & 99.9\(^{*}\) & 16.33\(^{*}\)
        \\

        Qwen3.5-4B
        & 52.5 & 51.0 & 97.2 & 1.40
        & 46.3\(^{*}\) & 45.8\(^{*}\)
        & 99.0\(^{*}\) & 30.55\(^{*}\)
        \\

        Qwen3.5-9B
        & 48.3 & 22.9 & 47.5 & 1.31
        & 71.7\(^{*}\) & 71.5\(^{*}\)
        & 99.7\(^{*}\) & 46.87\(^{*}\)
        \\

        VideoLLaMA3-7B
        & -- & 0.0 & 0.0 & --
        & -- & 0.0 & 0.0 & --
        \\

        InternVL3.5-8B
        & 1.1 & 1.1 & 100.0 & 9.72
        & 25.3\(^{*}\) & 25.1\(^{*}\)
        & 99.1\(^{*}\) & 311.92\(^{*}\)
        \\

        Molmo2-8B
        & 31.9 & 31.9 & 100.0 & 2.36
        & 12.5 & 11.1 & 89.1 & 247.15
        \\

        \midrule

        \textbf{BasketballSkills}
        & 31.2 & 30.5 & 97.7 & 1.89
        & 66.4 & 65.0 & 97.8 & 21.73
        \\

        \bottomrule
    \end{tabular*}

    \caption{
        Detailed Temporal Localization QA (Q7) results using the corrected
        temporal annotations.
        P-Acc@1s denotes accuracy over parseable outputs, while Acc@1s
        is computed over all examples, treating an unparseable output as
        incorrect.
        Parse denotes the output parse rate, and MAE is measured in seconds.
        For game-clock predictions, Acc@1s additionally requires the predicted
        period to match the reference period; game-clock MAE is computed only
        over parsed predictions with the correct period.
        \(^{*}\) indicates that the reported value differs from strict
        evaluation at the displayed precision after model-agnostic
        Period-Alias Normalization
        (\texttt{1ST}/\texttt{2ND}/\texttt{3RD}/\texttt{4TH} to
        \texttt{1}/\texttt{2}/\texttt{3}/\texttt{4}). The primary Q7 score in Table~2 is the macro-average of the two all-example Acc@1s columns. For VideoLLaMA3-7B, all Q7 outputs are schema-invalid and therefore cannot be parsed. Its all-example Acc@1s is reported as 0.0, while parsed accuracy and MAE are undefined and denoted by ``--''.
    }

    \label{tab:app-action-details}
\end{table}

Table~\ref{tab:app-action-details} reports the metrics for Temporal Localization QA under two
settings: video timestamp and game clock. The former requires the model
to output the video timestamp at which the event occurs, whereas the
latter requires it to output the time displayed on the scorebug's game
clock when the event occurs.

Period aliases account for a non-trivial share of nominal game-clock
parsing failures. Period-Alias Normalization recovers 84 of 790
Qwen2.5-VL-7B outputs, raising its Parse Rate from 89.2\% to 99.9\%
and its all-example Acc@1s from 39.1\% to 44.9\%. It additionally
recovers 5 Claude Sonnet 5, 53 Qwen3.5-4B, 7 Qwen3.5-9B, and 78
InternVL3.5-8B outputs. Because the repair uses only a fixed
equivalence between period-label spellings and never accesses the
ground-truth time, the recovered scores more faithfully measure
temporal localization rather than format compliance. All values that
differ from strict evaluation at the reported precision are marked
with \(^{*}\).

Qwen3.5-9B exhibits a distinct format-compliance failure in the
video-timestamp setting. Although all 790 requests receive a response,
415 predictions specify a temporal interval, such as
\([4.0, 4.5]\), rather than the requested single timestamp.
Consequently, only 47.5\% of its outputs are parseable, reducing its
all-example Acc@1s to 22.9\%. As a diagnostic analysis, among the 415
interval-valued predictions, selecting the interval start, midpoint,
or end makes 143, 165, or 174 of them correct, respectively. Combined
with the 181 correct point-valued predictions, these choices yield
all-example Acc@1s values of 41.0\%, 43.8\%, and 44.9\%,
respectively. Treating an interval as correct whenever it overlaps the
reference timestamp within the one-second tolerance gives 50.5\%,
close to Qwen3.5-4B's 51.0\%. However, because no unique point-valued
conversion is implied by an interval, we do not repair these outputs
in the reported results. VideoLLaMA3-7B produced no parseable Q7 outputs. It is therefore included in Table C.3 with an all-example Acc@1s of 0.0, while parsed-output metrics and further format-error diagnostics are unavailable.

\subsubsection{Play Event QA (Q8)}
\label{app:result-event-generation}

\begin{table}[!htbp]
\centering
\footnotesize
\setlength{\tabcolsep}{3.0pt}
\renewcommand{\arraystretch}{1.08}

\begin{tabular*}{\textwidth}{
    @{\extracolsep{\fill}}
    l
    rrr
    rrr
    rr
    @{}
}
    \toprule

    \multirow{2}{*}{\textbf{Model}}
    & \multicolumn{3}{c}{\textbf{Type+Participant}}
    & \multicolumn{3}{c}{\textbf{Type}}
    & \multirow{2}{*}{\textbf{PA}}
    & \multirow{2}{*}{\textbf{Parse}}
    \\

    \cmidrule(lr){2-4}
    \cmidrule(lr){5-7}

    & \textbf{P}
    & \textbf{R}
    & \textbf{F1}
    & \textbf{P}
    & \textbf{R}
    & \textbf{F1}
    & {}
    & {}
    \\

    \midrule

    GPT-5.4
    & 26.1 & 34.8 & 29.8
    & 52.2 & 69.5 & 59.6
    & 46.7 & 99.9
    \\

    Claude Sonnet 5
    & 11.3 & 19.5 & 14.3
    & 34.6 & 60.0 & 43.9
    & 24.5 & 95.5
    \\

    Gemini 3.5 Flash\(^{*}\)
    & 22.5 & 33.7 & 27.0
    & 43.2 & 64.7 & 51.8
    & 49.4 & 97.8
    \\

    \midrule

    Qwen2.5-VL-7B\(^{*}\)
    & 1.3 & 2.1 & 1.6
    & 22.7 & 34.9 & 27.5
    & 5.1 & 100.0
    \\

    Qwen3.5-4B\(^{*}\)
    & 6.5 & 11.1 & 8.2
    & 30.9 & 53.0 & 39.1
    & 20.6 & 100.0
    \\

    Qwen3.5-9B\(^{*}\)
    & 8.0 & 14.8 & 10.4
    & 35.3 & 65.2 & 45.8
    & 22.2 & 99.9
    \\

    VideoLLaMA3-7B
    & 0.0 & 0.0 & 0.0
    & 0.0 & 0.0 & 0.0
    & -- & 0.0
    \\

    InternVL3.5-8B\(^{*}\)
    & 0.3 & 1.6 & 0.5
    & 6.6 & 33.8 & 11.0
    & 2.7 & 100.0
    \\

    Molmo2-8B\(^{*}\)
    & 0.2 & 0.1 & 0.1
    & 10.4 & 3.0 & 4.7
    & 1.9 & 100.0
    \\

    \midrule

    \textbf{BasketballSkills}\(^{*}\)
    & \textbf{43.3} & \textbf{48.9} & \textbf{45.9}
    & \textbf{57.4} & \textbf{64.9} & \textbf{60.9}
    & \textbf{70.6} & \textbf{100.0}
    \\

    \bottomrule
\end{tabular*}

\caption{
    Detailed Play Event QA (Q8) results after Conservative JSON
    Recovery (\%).
    Under \emph{Type + Participant}, a match requires both the
    event type and the participant to be correct;
    under \emph{Type}, only the event type must match.
    P, R, and F1 denote LCS-based micro-averaged precision, recall,
    and F1, respectively.
    PA denotes Participant Accuracy over event-type-aligned event
    pairs, and Parse denotes the post-recovery schema-validity rate.
    \(^{*}\) indicates that Conservative JSON Recovery changes at
    least one output from the corresponding model. All VideoLLaMA3-7B outputs fail the required Q8 output schema.
Accordingly, its all-example sequence metrics and Parse Rate are
reported as 0.0. PA is undefined because no event-type-aligned
prediction--reference pairs can be formed.
}

\label{tab:app-event-details}
\end{table}

Table~\ref{tab:app-event-details} presents detailed metrics for Play Event QA, showing that BasketballSkills substantially outperforms MLLMs on all key metrics except event-type recall. MLLMs achieve markedly higher precision, recall, and F1 scores when required to predict only event labels than when required to jointly identify events and their associated players. Without player identification, GPT achieves 98\% of BasketballSkills’ F1 score; when both events and players must be identified, this proportion drops to only 65\%. These results highlight event-to-player association as a major weakness of current MLLMs.

Conservative JSON Recovery has its largest effect on the Qwen models. It
recovers 11, 189, and 366 outputs for Qwen2.5-VL-7B, Qwen3.5-4B, and
Qwen3.5-9B, respectively, increasing their Parse Rates from 98.9\%, 81.1\%,
and 63.3\% to 100.0\%, 100.0\%, and 99.9\%. Their repaired all-example
F1 scores are 1.6\%, 8.2\%, and 10.4\%, respectively. Recovery does
not necessarily increase F1: making a previously invalid prediction visible
to the evaluator can add both matched and unmatched events. The repair is
therefore used to expose the model's predicted content, rather than to
optimize its score; affected model names are marked with \(^{*}\).

\FloatBarrier

\end{document}